\documentclass[sigconf]{acmart}
\usepackage{multirow}
\usepackage{booktabs}
\usepackage{threeparttable}
\usepackage{algorithm}
\usepackage{enumitem}
\usepackage{algorithmic}
\usepackage{makecell}
\usepackage{amsthm,amsmath,mathrsfs}
\usepackage{amsfonts}
\usepackage{subfigure}

\newlength\savewidth
\newcommand\shline{\noalign{\global\savewidth\arrayrulewidth
                            \global\arrayrulewidth 1.5pt}%
                   \hline
                   \noalign{\global\arrayrulewidth\savewidth}
                   }
\usepackage{color, colortbl}
\definecolor{Gray}{gray}{0.9}
\usepackage{newfloat}
\usepackage{listings}
\DeclareCaptionStyle{ruled}{labelfont=normalfont,labelsep=colon,strut=off} % DO NOT CHANGE THIS
\floatstyle{ruled}
\newfloat{listing}{tb}{lst}{}
\floatname{listing}{Listing}

\AtBeginDocument{%
  \providecommand\BibTeX{{%
    \normalfont B\kern-0.5em{\scshape i\kern-0.25em b}\kern-0.8em\TeX}}}

\copyrightyear{2024}
\acmYear{2024}
\setcopyright{rightsretained}
\acmConference[SIGSPATIAL '24]{The 32nd ACM International Conference on Advances in Geographic Information Systems}{October 29-November 1, 2024}{Atlanta, GA, USA}
\acmBooktitle{The 32nd International Conference on Advances in Geographic Information Systems (SIGSPATIAL '24), October 29-November 1, 2024, Atlanta, GA, USA}
\acmDOI{10.1145/3678717.3691235}
\acmISBN{979-8-4007-1107-7/24/10}

\makeatletter
\gdef\@copyrightpermission{
  \begin{minipage}{0.3\columnwidth}
   \href{https://creativecommons.org/licenses/by/4.0/}{\includegraphics[width=0.90\textwidth]{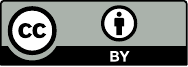}}
  \end{minipage}\hfill
  \begin{minipage}{0.7\columnwidth}
   \href{https://creativecommons.org/licenses/by/4.0/}{This work is licensed under a Creative Commons Attribution International 4.0 License.}
  \end{minipage}
  \vspace{5pt}
}
\makeatother

\begin{document}

%%
%% The "title" command has an optional parameter,
%% allowing the author to define a "short title" to be used in page headers.
\title{Towards Unifying Diffusion Models for Probabilistic Spatio-Temporal Graph Learning}

%%
%% The "author" command and its associated commands are used to define
%% the authors and their affiliations.
%% Of note is the shared affiliation of the first two authors, and the
%% "authornote" and "authornotemark" commands
%% used to denote shared contribution to the research.
% \author{Junfeng Hu$^1$, Xu Liu$^1$, Zhencheng Fan$^2$, Yuxuan Liang$^3$, Roger Zimmermann$^1$}
% \affiliation{%
%   \institution{$^1$National University of Singapore, $^2$University of Technology Sydney, \\ $^3$The Hong Kong University of Science and Technology (Guangzhou)}
%   % \country{Singapore}
% }
% \email{{junfengh, liuxu, rogerz}@comp.nus.edu.sg; zhencheng.fan@student.uts.edu.au; yuxliang@outlook.com}
% \authornote{Corresponding author.}

\author{Junfeng Hu}
\orcid{0000-0003-1409-1495}
\affiliation{%
  \institution{National University of Singapore}
  \country{Singapore}
}
\email{junfengh@u.nus.edu}

\author{Xu Liu}
\orcid{0000-0003-2708-0584}
\affiliation{%
  \institution{National University of Singapore}
  \country{Singapore}
}
\email{liuxu@comp.nus.edu.sg}

\author{Zhencheng Fan}
\orcid{0000-0002-8046-2541}
\affiliation{%
  \institution{University of Technology Sydney}
  \city{Sydney}
  \state{NSW}
  \country{Australia}
}
\email{zhencheng.fan@student.uts.edu.au}

\author{Yuxuan Liang}
\orcid{0000-0003-2817-7337}
\affiliation{%
  \institution{Hong Kong University of Science and Technology (Guangzhou)}
  \country{China}
}
\email{yuxliang@outlook.com}
\authornote{Corresponding author. The code is available at: \url{https://github.com/hjf1997/USTD}.}

\author{Roger Zimmermann}
\orcid{0000-0002-7410-2590}
\affiliation{%
  \institution{National University of Singapore}
  \country{Singapore}
}
\email{rogerz@comp.nus.edu.sg}

%%
%% By default, the full list of authors will be used in the page
%% headers. Often, this list is too long, and will overlap
%% other information printed in the page headers. This command allows
%% the author to define a more concise list
%% of authors' names for this purpose.
\renewcommand{\shortauthors}{Junfeng Hu et al.}

%%
%% The abstract is a short summary of the work to be presented in the
%% article.
\begin{abstract}
Spatio-temporal graph learning is a fundamental problem in modern urban systems.
Existing approaches tackle different tasks independently, tailoring their models to unique task characteristics. These methods, however, fall short of modeling intrinsic uncertainties in the spatio-temporal data. 
Meanwhile, their specialized designs misalign with the current research efforts toward unifying spatio-temporal graph learning solutions. 
In this paper, we propose to model these tasks in a unified probabilistic perspective, viewing them as predictions based on conditional information with shared dependencies. Based on this proposal, we introduce Unified Spatio-Temporal Diffusion Models (USTD) to address the tasks uniformly under the uncertainty-aware diffusion framework.
USTD is holistically designed, comprising a shared spatio-temporal encoder and attention-based denoising decoders that are task-specific.
The encoder, optimized by pre-training strategies, effectively captures conditional spatio-temporal patterns.
The decoders, utilizing attention mechanisms, generate predictions by leveraging learned patterns.
Opting for forecasting and kriging, the decoders are designed as Spatial Gated Attention (SGA) and Temporal Gated Attention (TGA) for each task, with different emphases on the spatial and temporal dimensions.
Combining the advantages of deterministic encoders and probabilistic decoders, USTD achieves state-of-the-art performances compared to both deterministic and probabilistic baselines, while also providing valuable uncertainty estimates.
\end{abstract}

%%
%% The code below is generated by the tool at http://dl.acm.org/ccs.cfm.
%% Please copy and paste the code instead of the example below.
%%


% \ccsdesc[500]{Do Not Use This Code~Generate the Correct Terms for Your Paper}
% \ccsdesc[300]{Do Not Use This Code~Generate the Correct Terms for Your Paper}
% \ccsdesc{Do Not Use This Code~Generate the Correct Terms for Your Paper}
% \ccsdesc[100]{Do Not Use This Code~Generate the Correct Terms for Your Paper}

%%
%% Keywords. The author(s) should pick words that accurately describe
%% the work being presented. Separate the keywords with commas.

%% A "teaser" image appears between the author and affiliation
%% information and the body of the document, and typically spans the
%% page.

%%
%% This command processes the author and affiliation and title
%% information and builds the first part of the formatted document.
\maketitle

\section{Introduction}
\begin{figure}
  \centering
  \includegraphics[width=0.99 \linewidth]{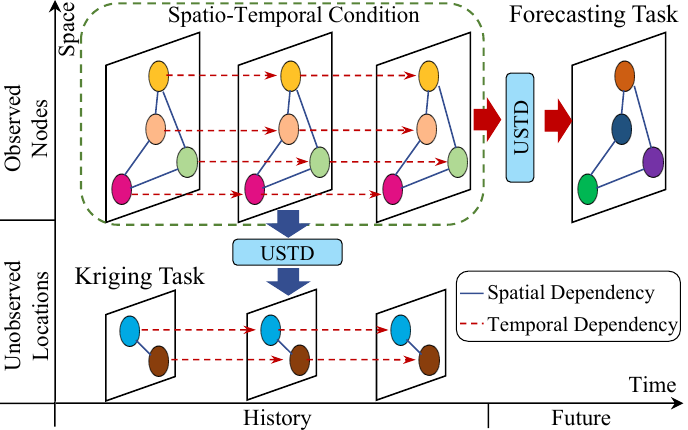}
  \vspace{-0.5em}
  \caption{Spatio-temporal graph learning tasks involve modeling conditional distributions based on the same conditional information with complex spatio-temporal patterns.}
    \label{fig:intro}
    \vspace{-1em}
\end{figure}

Spatio-temporal graph data, characterized by inherent spatial and temporal patterns, is ubiquitous in modern urban systems.
The learning of graph data, through analyzing its spatial and temporal patterns, enables multiple downstream tasks such as spatio-temporal \emph{forecasting} and \emph{kriging}. The forecasting task aims to predict future trends of specific locations based on their historical conditions. Kriging, on the other hand, requires estimating the states of unobserved locations using data from observed ones during the same period.
These tasks have practical impacts on inferring future knowledge and compensate for data sparsity, which facilitates a myriad of real-world applications such as smart cities~\cite{bermudez2018spatio,trirat2021df,wang2020traffic}, climate analysis~\cite{schweizer2022semi,tempelmeier2019data4urbanmobility}, and human mobility modeling~\cite{luo2021stan,zhang2023automated}.

Existing efforts tackle spatio-temporal graph learning problems separately by introducing dedicated models that account for the characteristics of each task. For instance, Spatio-Temporal Graph Neural Networks (STGNNs) have emerged as a favorite for forecasting to model dependencies in historical data~\cite{li2018dcrnn_traffic,zheng2020gman}. 
These methods typically rely on sequential models for temporal capturing and graph neural networks for spatial modeling.
Kriging methods, also leveraging STGNNs to capture dependencies within the observed data, place a greater emphasis on the structural correlations between observed data and unobserved locations, which are captured by various graph aggregators~\cite{wu2021spatial,hu2023graph}. 

Despite the notable achievements of these methods, their deterministic nature falls short of modeling data uncertainties, limiting their reliability as trust-worthy solutions. Moreover, the specialized designs are tailored for individual tasks only, costing high deployment expenses for varied application scenarios and misaligning with the current research trends toward unifying task solutions~\cite{jin2023spatio}.
These concerns motivate us to explore the possibility of a unified model design for uncertainty-aware spatio-temporal graph learning. In this work, we argue that the learning tasks can be summarized as modeling the conditional distribution $P_{\phi,\theta}(Y|X)$, where the dependencies in condition $X$ are shared among tasks (as shown in Fig.~\ref{fig:intro}).
To achieve this, an intuitive idea is to first leverage a shared network, parameterized by $\phi$, to extract deterministic conditional patterns. Then, task-specific probabilistic models can be utilized to learn the respective distributions and obtain the predictions $Y$. Among the recent models, we propose to utilize diffusion probabilistic modeling
(DDPM)~\cite{ho2020denoising} to parameterize $\theta$. Known for effectively modeling complex distributions, DDPM is an underexplored but promising framework for spatio-temporal graph learning~\cite{huang2022generative}. 

However, it is non-trivial to learn conditional distributions by DDPM due to the following challenges. 
(i) \emph{How to learn deterministic spatio-temporal patterns in the conditional information $X$.} 
Conditional patterns are deterministic and should be pre-extracted to serve as input for probabilistic models.
Existing methods, however, model them by encoders trained alongside DDPM's denoising networks, which leads to increased optimization difficulties~\cite{rasul2021autoregressive,alcaraz2023diffusionbased}. The poorly captured dependencies impede DDPM from outperforming deterministic models~\cite{wen2023diffstg}.
(ii) \emph{How to learn the distribution, based on extracted conditional patterns, to generate targets $Y$}. Generating targets for different tasks entails various emphases on the temporal and spatial dependencies. Unfortunately, existing diffusion models treat the tasks simply as a reconstruction problem, where the denoising networks process the target variables and conditions indiscriminately, thereby neglecting such distinctions~\cite{tashiro2021csdi,liu2023pristi}. 
Thus, unsatisfactory performances are obtained on all the tasks.

To tackle these challenges, we propose a diffusion-based framework for spatio-temporal graph learning, termed Unified Spatio-Temporal Diffusion (USTD). To solve the first challenge, we propose a spatio-temporal encoder that is shared by all downstream tasks. The encoder is pre-trained using an unsupervised autoencoding mechanism, with a graph sampling method and a masking strategy~\cite{he2022masked} to enhance its capability of capturing conditional dependencies.
Specifically, the encoder, consisting of spatio-temporal layers, maps the conditional data to representations in low-dimensional latent space~\cite{hinton1993autoencoders}. Then, a lightweight decoder is adopted to reconstruct the data. In this way, the denoising modules, taking in representations with rich dependencies learned by the encoder, only focus on data prediction, which alleviates optimization burdens.

Regarding the second challenge, we propose the use of denoising networks to learn conditional distributions and generate predictions. As forecasting and kriging entail discrepant dependency emphasis, we introduce Temporal Gate Attention (TGA) and Spatial Gated Attention (SGA) for them respectively, with each having unique attention mechanisms to model relations between the predictions and representations. The TGA captures correlations on the temporal dimension, while the SGA learns dependencies on the spatial dimension. 
Through this approach, the networks exclusively concentrate on predicting data by utilizing relations on the most crucial dimension, leading to improved efficiency. Meanwhile, dependencies from the other dimension are still learned by the encoder, which guarantees the performance of our model.
In summary, our contributions lie in three aspects:
\begin{itemize}[leftmargin=*]
    \item We take an early step towards unifying two spatio-temporal graph learning tasks -- forecasting and kriging -- into a diffusion framework with uncertainty estimates.
    \item We introduce an encoder to effectively capture shared conditional patterns and propose attention-based denoising networks to generate predictions efficiently.
    % here needs to change, to emphasize the contribution of TGA, SGA
    \item Extensive experiments are conducted to evaluate the performance of USTD. The results consistently demonstrate our model's superiority over baselines on both tasks, with a maximum reduction of 12.0\% in Continuous Ranked Probability Score and 4.9\% in Mean Average Error.
\end{itemize}

\section{Preliminaries}
In this section, we first define the notations of spatio-temporal graph data and the forecasting and kriging problems. Subsequently, we provide a brief introduction to DDPM.

\subsection{Problem Formulation and Notations}

\emph{Definition 1 (Spatio-Temporal Graph).} A graph is represented as $\mathcal{G}=(\mathcal{V}, \mathcal{E})$, where $\mathcal{V}$ is the node set with $|V|=N$ and $\mathcal{E}$ is the edge set. Based on $\mathcal{E}$, the adjacency matrix $A$ is calculated to measure the non-Euclidean distances between neighboring nodes. For each node $i$, we denote its signals over a time window $T$ as $X_i=(x_i^1, .., x_i^t, .., x_i^T)\in \mathbb{R}^{T\times d_x}$, where $d_x$ is the number of channels. We denote $X=(X_1, .., X_N)^\top=(X^1, .., X^T) \in \mathbb{R}^{N\times T \times d_x}$ as the data of all $N$ nodes over the window $T$. Appendix \ref{app:notation} summarizes the main notations in this paper.
\vspace{0.3em}

\noindent\emph{Definition 2 (Forecasting).} The forecasting problem aims to learn a function $f(\cdot)$ that predicts future signals of all $N$ nodes over $T^\prime$ steps given their historical data from the past $T$ steps. 
\begin{equation}
    X^{1:T} \stackrel{f}{\longrightarrow} Y^{T+1: T+T^\prime},
\end{equation}
where $Y^{T+1: T+T^\prime}\in \mathbb{R}^{N\times T^\prime \times d_y}$ is the future readings. 
\vspace{0.3em}

\noindent\emph{Definition 3 (Kriging).} The goal of kriging is to learn a function $f(\cdot)$ that predicts signals of $M$ unobserved locations based on the $N$ observed nodes over the same time period $T$.
\begin{equation}
    X_{1:N} \stackrel{f}{\longrightarrow} Y_{N+1:N+M},
\end{equation}
where $Y_{N+1:N+M}\in \mathbb{R}^{M\times T \times d_y}$ is signals of unobserved locations.

\subsection{Denoising Diffusion Probabilistic Models}
Denoising diffusion probabilistic model (DDPM)~\cite{ho2020denoising,yang2024survey} generates target samples by learning a distribution $p_\theta(x_0)$ that approximates the target distribution $q(x_0)$. DDPM is a latent model that introduces additional latent variables $(x_1, .., x_K)$\footnote{We use subscript $k$ to index diffusion steps, distinguishing it from node indicator $i$.} in a way the marginal distribution is defined as $p_\theta\left(x_0\right)=\int p_\theta\left(x_{0:K}\right) \mathrm{d} x_{1:K}$. Inside, the joint distribution, assuming conditional independence, is decomposed into a Markov chain, resulting in a learning procedure involving forward and reverse processes.

The forward process contains no learnable parameters and is a combination of Gaussian distributions following the chain:
\begin{equation}
    q\left(x_{1:K} \mid x_0\right)=\prod_{k=1}^K \underbrace{q\left(x_k \mid x_{k-1}\right)}_{\mathcal{N}\left(x_k ; \sqrt{1-\beta_k} x_{k-1}, \beta_k \boldsymbol{I}\right)},
\label{eq:ddpm_forward}
\end{equation}
where $\beta_k\in(0, 1)$ is an increasing variance hyperparameter representing the noise level. Instead of sampling $x_k$ step by step following the chain, Gaussian distribution entails a one-stop sampling manner written as $q(x_k|x_0)=\mathcal{N}\left(x_k ; \sqrt{\alpha_k} x_0,\left(1-\alpha_k\right) \boldsymbol{I}\right)$, where $\alpha_k=\prod_{s=1}^k \hat{\alpha}_s$ and $\hat{\alpha}_k=1-\beta_k$. Therefore, $x_k$ can be sampled through the expression $x_k=\sqrt{\alpha_k} x_0+\sqrt{1-\alpha_k} \epsilon$ with $\epsilon \sim \mathcal{N}(0, \boldsymbol{I})$. 
Then, in the reverse process, a network denoises $x_k$ to recover $x_0$ following the reversed chain. Starting from a sampled Gaussian noise $x_K \sim \mathcal{N}\left(0, \boldsymbol{I}\right)$, this process is formally defined as:
\begin{equation}
    p_\theta\left(x_{0:K}\right)=p(x_K) \prod_{k=1}^K \underbrace{p_\theta\left(x_{k-1} \mid x_k\right)}_{\mathcal{N}\left(x_{k-1} ; \mu_\theta\left(x_k, k\right), \sigma_\theta\left(x_k, k\right) \boldsymbol{I}\right)},
\end{equation}
where $\mu_\theta(\cdot)$ and $\sigma_\theta(\cdot)$ are denoising networks. The model can be optimized by minimizing the negative evidence lower-bound (ELBO):
\begin{equation}
\mathcal{L}(\theta) = \mathbb{E}_{q\left(x_{0:K}\right)}\left[-\log p\left(x_K\right)-\sum_{k=1}^K \log \frac{p_\theta\left(x_{k-1} \mid x_k\right)}{q\left(x_k \mid x_{k-1}\right)}\right].
\end{equation}
DDPM~\cite{ho2020denoising} suggests it can be trained efficiently by a simplified parameterization schema, which leads to the following objective:
\begin{equation}
\mathcal{L}(\theta)= \mathbb{E}_{x_0,\epsilon,k}\left\|\epsilon-\epsilon_\theta\left(\sqrt{\alpha_k} x_0+\sqrt{1-\alpha_k} \epsilon, k\right)\right\|_2^2,
\label{eq:ddpmloss}
\end{equation}
where $\epsilon_\theta(\cdot)$ is a network estimating noise added to $x_k$.
Once trained, target variables are first sampled from Gaussian as the input of $\epsilon_\theta(\cdot)$ to progressively learn the distribution $p_\theta(x_{k-1}|x_k)$ and denoise $x_k$ until $x_0$ is obtained. DDPM decomposes a distribution into a combination of Gaussian, with each step only recovering a simple Gaussian. This capability empowers the model to effectively represent complex distributions, making it suitable for learning the conditional distributions in our tasks.

\begin{figure}[!h]
  \centering
  \includegraphics[width=1 \linewidth]{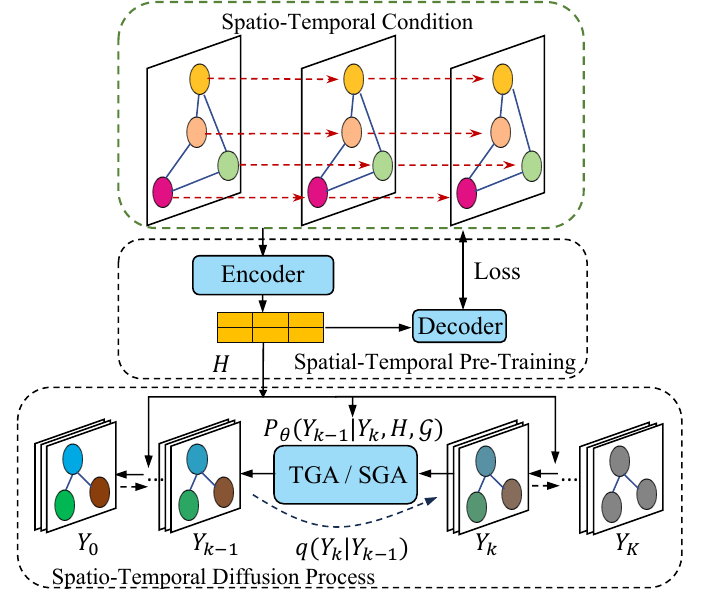}
  \caption{The USTD framework comprises a pre-trained spatio-temporal encoder and attention-based denoising networks TGA and SGA for the diffusion process. $H$ is the conditional representation learned by the encoder.}
    \label{fig:framework}
\end{figure}

\section{Unified Spatio-Temporal Diffusion Models}
Fig.~\ref{fig:framework} shows the overall pipeline of USTD, which consists of two major components: a pre-trained encoder and task-specific denoising decoders for diffusion processes. The encoder aims to learn high-quality representations of conditional information while the denoising networks in the diffusion processes generate predictions using the representations.

\subsection{Pre-Training Spatial-Temporal Encoder}
\begin{figure}
  \centering
  \includegraphics[width=1 \linewidth]{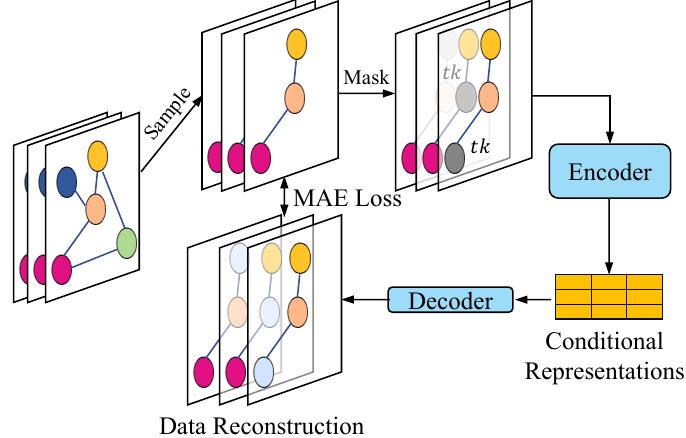}
  \caption{Pipeline of the spatio-temporal encoder, where $tk$ denotes the mask token.}
    \label{fig:encoder}
\end{figure}

Conditional information contains abundant spatio-temporal dependencies that play a vital role and can be shared in spatio-temporal tasks.
Moreover, it is deterministic and should be pre-extracted effectively for the benefit of subsequent denoising models.
% Based on this insight, we introduce a pre-trained encoder that is optimized through an unsupervised autoencoding strategy~\cite{hinton1993autoencoders}. 
Based on this insight, we propose to pre-train an encoder based on the unsupervised autoencoding strategy~\cite{hinton1993autoencoders}. 
As shown in Fig.~\ref{fig:encoder}, it employs the encoder to acquire conditional representations of given data in latent space, with an auxiliary decoder to reconstruct the data from the space. The learned latent space has demonstrated effectiveness in capturing data correlations, thereby facilitating downstream tasks~\cite{liu2023spatio,kipf2016variational}. Here, we first introduce the network architecture and then discuss the graph sampling and masking strategies to enhance the model's learning capability.

\paragraph{Encoder} The encoder consists of a stack of STGNN layers, with each capturing spatial and temporal correlations. In each layer, we first employ a temporal convolution network (TCN) for temporal dependencies. Specifically, a gated 1D convolution~\cite{wu2019graph} updates conditional representations $H_i\in\mathbb{R}^{T \times d_h}$ from the last layer as follows:
\begin{equation} 
    H_i = g(H_i \star \mathcal{K}_1) \odot \sigma(H_i \star \mathcal{K}_2),
\end{equation}
where $\star\mathcal{K}_1$ and $\star\mathcal{K}_2$ are convolutions with a kernel size of $c\times d_h\times d_{out}$, $\odot$ is the Hadamard product, $g(\cdot)$ is an activation function and $\sigma(\cdot)$ is the sigmoid function. Then, a graph convolution network (GCN)~\cite{thomas2017semi} is utilized to capture spatial relations, which can be formulated as:
\begin{equation}
    H = \sum_{l=0}^L A^l H W_l,
\end{equation}
where $W_l\in \mathbb{R}^{d_h\times d_{out}}$ is a learnable matrix and $L$ is the depth of graph propagation. Meanwhile, skip and residual connections are added to transfer information over different layers. 

To obtain conditional representations in the latent space, the encoder takes in conditional information $X\in\mathbb{R}^{N\times T\times d_x}$ and yields the representations $H\in\mathbb{R}^{N\times \tau\times d_h}$. Note that we do not use zero padding for TCNs so the temporal dimension is squashed with $\tau << T$. In this way, the condition is mapped into low-dimensional latent space containing representative information. 
Moreover, the denoising network, taking the representations as input, requires less computation, which decreases diffusion's well-known prolonged sampling time~\cite{tashiro2021csdi,liu2023pristi}.

\paragraph{Decoder}
The decoder is a lightweight network to reconstruct the original conditional data from the learned representations. It contains a small stack of spatio-temporal blocks introduced above that takes in the representations $H\in\mathbb{R}^{N\times \tau\times d_h}$ and outputs $H\in\mathbb{R}^{N\times 1 \times d_h}$. Then, a multi-layer perceptron layer is used to obtain the reconstruction $\hat{X}\in\mathbb{R}^{N\times T\times d_x}$. The encoder and decoder are optimized by minimizing the mean absolute error loss between the ground truth $X$ and the reconstruction $\hat{X}$. 

\paragraph{Graph Sampling}
% The pre-training and kriging tasks involve feeding the encoder with different graph structures, requiring it to generalize across various graphs~\cite{wu2021spatial}.
The encoder only processes graph structures of observed nodes, resulting in different input graphs for pre-training and kriging tasks.
To enable the encoder to generalize across various graphs, we employ a simple graph sampling mechanism. For each iteration, we sample a subset of nodes $\bar{\mathcal{V}} \in \mathcal{V}$. Then, the new graph $\bar{\mathcal{G}}$ and its adjacency matrix $\bar{A}$ are constructed. The training objective is altered to reconstruct signals of nodes in the sampled graph. In this way, the model is unable to capture spatio-temporal relations by memorizing the graph structure, increasing its generalizability.

\paragraph{Masking} 
The dimension $d_h$ of the latent space is much larger than the node channel $d_x$; thus, the model risks learning a trivial solution that identically maps node data into the space~\cite{grill2020bootstrap}. To alleviate the problem, we utilize a masking strategy that randomly corrupts the conditional information during training~\cite{he2022masked}. We sample a binary mask $MSK\in\{0,1\}^{N\times T}$, where $1$ indicates data to be corrupted. Then, the data is masked with a learnable token $tk\in\mathbb{R}^{d_x}$, which can be defined as:
\begin{equation}
    \bar{x}_i^t = \begin{cases} x_i^t & MSK_i^t=0 \\ tk & MSK_i^t = 1\end{cases},
\end{equation}
where $\bar{x}_i^t$ is the masked data. Accordingly, the loss is changed to measure the masked signals of nodes in the sampled graph, given the partially observed node data $\bar{X}$ and the matrix $\bar{A}$.
Following MAE~\cite{he2022masked}, we use a high masking ratio of 75\%, forcing the model to extract dependencies from scarce signals and learn meaningful node representations.

\subsection{Spatio-Temporal Diffusion Process} 
Spatio-temporal graph learning tasks can be viewed as modeling conditional distributions given learned representations. 
To learn such distributions, we leverage denoising diffusion modeling~\cite{ho2020denoising}. We first introduce the formulation, training, and inference procedures for conditional DDPM in our tasks, followed by the descriptions of the task-specific denoising networks TGA and SGA.

\paragraph{Conditional Diffusion Formulation}
Given the prediction target $Y\in \mathbb{R}^{N\times T^\prime \times d_y}$ for forecasting or $Y\in \mathbb{R}^{M\times T\times d_y}$ for kriging, the diffusion process first transforms it into a sequence of variables $(Y_0, .., Y_k, .., Y_K)$ with $Y_0=Y$. Specifically, Gaussian noises are gradually added to $Y_k$ such that $Y_K$ is a standard Gaussian, which is the same as DDPM's forward process in Eq.~\ref{eq:ddpm_forward}. 
Then, the reverse process employs a denoising network to reconstruct the target distribution based on conditional representations and the graph structure. The process samples a target from a standard Gaussian at the first step and then progressively predicts a less noisy one, learning to the following expression:
\begin{equation}
    p_\theta\left(Y_{0: K} | H, \mathcal{G} \right)=p\left(Y_K\right) \prod_{k=1}^K p_\theta\left(Y_{k-1} \mid Y_k, H, \mathcal{G}\right),
\end{equation}
where $\theta$ is the parameters of the denoising network.

\paragraph{Training} 
The model can be optimized by variational approximation with the simplified loss in Eq.~\ref{eq:ddpmloss} as follows:
\begin{equation}
    \mathcal{L}(\theta)= - \mathbb{E}_{Y_0,\epsilon,k}\left\|\epsilon-\epsilon_\theta\left(\sqrt{\alpha_k} Y_0+\sqrt{1-\alpha_k} \epsilon, H, \mathcal{G}, k\right)\right\|_2^2,
\end{equation}
where $\epsilon\sim\mathcal{N}\left(0, \boldsymbol{I}\right)$ is a random noise and $\epsilon_\theta(\cdot)$ is the denoising network described in the following section. The complete training procedure is summarized in Algo.~\ref{algo:train}.

\begin{algorithm}[h]
\caption{Training for denoising networks}
\begin{algorithmic}[1]
\REQUIRE Distribution of training data $q(Y)$, graph $\mathcal{G}$, pre-trained encoder $\operatorname{Enc}_\phi(\cdot)$
\ENSURE Trained denoising function $\epsilon_\theta(\cdot)$
\REPEAT
\STATE $Y_0\sim q(Y)$, $k\sim Uniform(1,..,K)$, $\epsilon\sim\mathcal{N}(0, \boldsymbol{I})$
\STATE Get conditional information $X$ from the dataset given $Y_0$, learn
representations $H$ from the encoder $\operatorname{Enc}_\phi(X, \mathcal{G})$
\STATE Take gradient descent step \\
 \quad $\nabla_\theta\left\|\epsilon-\epsilon_\theta\left(\sqrt{\alpha_k} Y_0+\sqrt{1-\alpha_k} \epsilon, H, \mathcal{G}, k\right)\right\|_2^2$
\UNTIL{Converged}
\end{algorithmic}
\label{algo:train}
\end{algorithm}

\paragraph{Inference}
During inference, we first extract the conditional representations $H$ by the pre-trained encoder. Then, conditioned on $H$, a standard Gaussian sample is denoised progressively along the reversed chain to predict the target. The process of each denoising step is formulated as follows: 
\begin{equation}
    Y_{k-1}=\frac{1}{\sqrt{\hat{\alpha}_k}}\left(Y_k-\frac{\beta_k}{\sqrt{1-\alpha_k}} \epsilon_\theta\left(Y_k, H, \mathcal{G}, k\right)\right)+\sqrt{\beta_k} Z,
\label{eq:inference}
\end{equation}
where $Z$ is a standard Gaussian noise. Algo.~\ref{algo:inference} describes the full procedure of the inference process and we provide detailed descriptions of the training and inference in Appendix~\ref{app:training}. 

\begin{algorithm}[h]
\caption{Sampling target $Y$}
\begin{algorithmic}[1]
\REQUIRE Conditional information $X$, graph $\mathcal{G}$, denoising network $\epsilon_\theta(\cdot)$, encoder $\operatorname{Enc}_\phi(\cdot)$  
\ENSURE Inference target $Y$
\STATE Get conditional representations $H$ from the pre-trained encoder $\operatorname{Enc}_\phi(X, \mathcal{G})$
\STATE $Y_K\sim\mathcal{N}(0, \boldsymbol{I})$, $Z\sim\mathcal{N}(0, \boldsymbol{I})$ if $k>1$ else $Z=0$
\FOR{$k=K$ to $1$}
\STATE $Y_{k-1}=\frac{1}{\sqrt{\hat{\alpha}_k}}\left(Y_k-\frac{\beta_k}{\sqrt{1-\alpha_k}} \epsilon_\theta\left(Y_k, H, \mathcal{G}, k\right)\right)+\sqrt{\beta_k} Z,$
\ENDFOR
\RETURN $Y_0$
\end{algorithmic}
\label{algo:inference}
\end{algorithm}

\begin{figure}
  \centering
  \includegraphics[width=1 \linewidth]{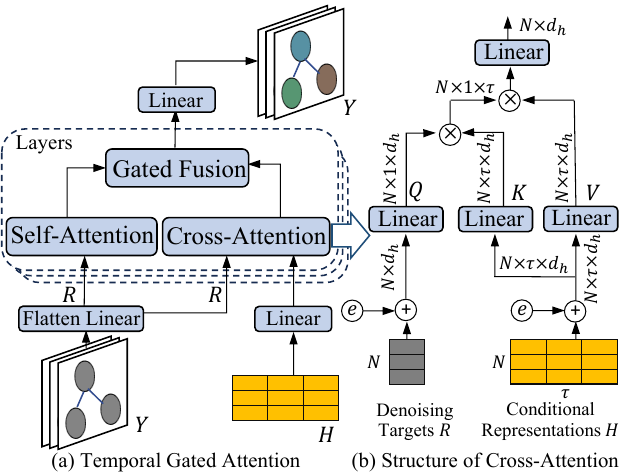}
  \caption{(a) Architecture of the proposed Temporal Gated Attention Network. (b) Pipeline of the cross-attention, where $e$ denotes time and diffusion embeddings.}
    \label{fig:tga}
\end{figure}

\paragraph{Temporal Gated Attention Network}
The denoising network is crucial in predicting the target data by capturing its dependencies with conditional representations. As forecasting predicts future signals based on historical data, with a focus on temporal dimension, we propose a temporal gated attention network (TGA) that predicts future data of a node by capturing dependencies with its historical representations. 
As shown in Fig.~\ref{fig:tga}(b), given the predicted target $Y\in\mathbb{R}^{N\times T^\prime \times d_y}$ from the previous diffusion step\footnote{We omit the subscript $k$ here for succinct.}, we first obtain its ﬂattened embeddings $R\in\mathbb{R}^{N\times d_h}$ by a linear layer. Then, a cross-attention block~\cite{vaswani2017attention} is used to measure the dependencies between $R_i\in\mathbb{R}^{d_h}$ of node $i$ and its historical representations $H_i\in\mathbb{R}^{\tau\times d_h}$:
\begin{align}
\begin{split}
    &Q_i = W_{cq} R_i, \quad K_i = W_{ck} H_i, \quad  V_i = W_{cv} H_i,\\
    & R^{ca}_i = \operatorname{softmax}\left(\frac{Q_i K_i^T}{\sqrt{d_h}}\right) V_i,
\end{split}
\end{align}
where $R_i^{ca}\in\mathbb{R}^{d_h}$ is node $i$'s outputs, $W_{cq}$, $W_{ck}$, $W_{cv}$ are learnable matrices. Note that the attention is performed independently for each node. 
Next, to capture the correlations among the node embeddings $R_i$, we introduce a self-attention block implementing $R^{sa}=\operatorname{softmax}\left(\frac{Q K^T}{\sqrt{d_h}}\right) V$ with
% \begin{equation}
    $Q = W_{sq} R$, $K = W_{sk} R$, and $V = W_{sv} R$,
% \end{equation}
where $R^{sa}\in\mathbb{R}^{N\times d_h}$ is the self-attention outputs.
Then, a gated fusion mechanism is proposed to integrate the outputs:

\begin{align}
\begin{split}
    &R= Gate \odot R^{ca} + (1-Gate) \odot R^{sa},\\
    &Gate=\sigma\left(R^{ca} W_{g1}+R^{sa} W_{g2}+b_g\right),
\end{split}
\end{align}
where $W_{g1}, W_{g1}, b_g$ are parameters and $\sigma(\cdot)$ is a sigmoid function. The gated attention can be stacked with several layers. In the end, we utilize a linear layer to get the prediction of the current diffusion step. In this way, the module only captures dependencies on the temporal dimension that are most important. Compared to attention operates on both spatial and temporal dimensions, the time complexity reduces from $\mathcal{O}((N\times (\tau + 1))^2)$ to $\mathcal{O}(N\times (\tau + 1 + N))$. Meanwhile, the spatial dependencies, involved in the representations, are still captured by the encoder, which guarantees the model's capability on spatial awareness. 

\paragraph{Spatial Gated Attention Network} For kriging, we similarly propose spatial gated attention (SGA) with two differences from TGA. First, we employ an embedding layer to absorb the temporal dimension of the conditional representations such that $H\in\mathbb{R}^{N\times d_h}$. Second, the cross-attention captures spatial dependencies between target nodes and representations of observed nodes, obtaining the outputs $R^{ca}\in\mathbb{R}^{M\times d_h}$. The rest parts remain the same as TGA's structure. 
Benefited by the design, the complexity reduced from $\mathcal{O}(((M+N) \times \tau)^2)$ to $\mathcal{O}((M + N)\times M)$.
Note that following the previous work~\cite{tashiro2021csdi}, we enhance TGA and SGA by incorporating diffusion, time, or space embeddings as additional learning contexts. Please refer to Appendix~\ref{app:ustd} for SGA's detailed descriptions.

\section{Experiments}
\subsection{Datasets}
We evaluate USTD on four real-world datasets from two spatio-temporal domains. Table~\ref{tab:dataset} provides a summary of the datasets and more details are described in Appendix~\ref{app:dataset}.
\begin{itemize}[leftmargin=*]
    \item \emph{Traffic}: \textbf{PEMS-03}~\cite{song2020spatial} provides measurements of traffic flow, offering information into the dynamic traffic conditions within the San Francisco Bay Area. \textbf{PEMS-BAY}~\cite{li2018dcrnn_traffic} contains traffic speed information from the same area. The adjacency matrices are constructed by road networks defined by the datasets.
    \item \emph{Air Quality}: \textbf{AIR-BJ} and \textbf{AIR-GZ} compile one-year data on air quality indexes (AQI) from monitoring stations in the cities of Beijing and Guangzhou, China. %In our experiments, we focus on the most critical index PM2.5.
    The datasets provide various quality attributes and we utilize the most critical index PM2.5. The adjacency matrices are calculated by the geological locations of stations.
\end{itemize}
%where more details are described in Appendix~\ref{app:dataset}.
\begin{table}[!h]
\caption{Statistics of the datasets. \# denotes the number of.}
% \vspace{-1.em}
  \scalebox{1}{
  \centering
  \begin{tabular}[width=0.78\linewidth]{l||cccc}
    \shline
       Dataset & \#Node & \#Time Step & Granularity & Attribute \\
       \hline
       \hline
       PEMS-03 & 358 & 26,208 & 5 min & Flow \\
       PEMS-BAY & 325 & 52,116 & 5 min & Speed  \\
       AIR-BJ & 36 &  8,760 & 1 hour & PM2.5 \\
       AIR-GZ & 42 & 8,760 & 1 hour & PM2.5 \\
  \shline
\end{tabular}
}
% \vspace{-1.em}
\label{tab:dataset}
\end{table}

\subsection{Baselines}
\begin{figure}
  \centering
  \includegraphics[width=1.06 \linewidth]{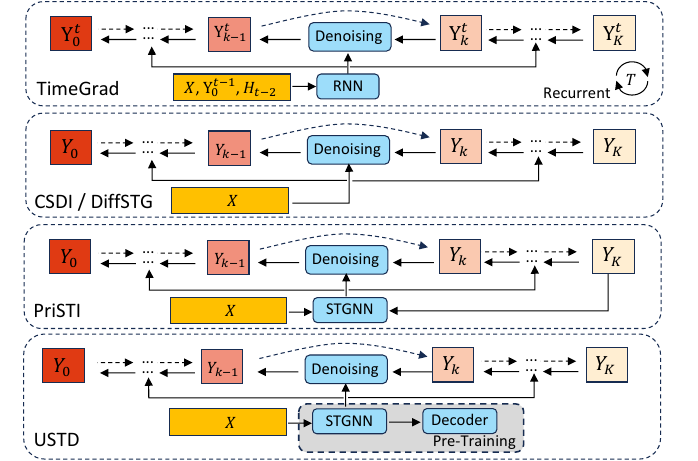}
  \caption{Comparison among diffusion models. Denoising means denoising networks and the yellow blocks represent conditional information.}
    \label{fig:comp}
\end{figure}

We consider the following 16 baselines for the two tasks.
\begin{itemize}[leftmargin=*]
    \item \emph{Forecasting}: 
    \textbf{STGCN}~\cite{yu2018stgcn}, \textbf{DCRNN}~\cite{li2018dcrnn_traffic}, \textbf{GWN}~\cite{wu2019graph}, \textbf{STSGCN}~\cite{song2020spatial}, \textbf{GMSDR}~\cite{liu2022msdr} are popular STGNNs that propose various mechanism to enhance the performance.
    \textbf{STGNCDE}~\cite{choi2022graph}, \textbf{STGODE}~\cite{fang2021spatial} are spatio-temporal networks empowered by neural differential equations.
    %\textbf{STGCN}~\cite{yu2018stgcn} and \textbf{DCRNN}~\cite{li2018dcrnn_traffic} are pioneer methods for traffic forecasting, which utilizes TCNs or RNNs for temporal learning, followed by GCNs for spatial dependencies.
    %\textbf{GWN}~\cite{wu2019graph} is an influential model that explicitly captures hidden spatial relations among nodes. \textbf{STSGCN}~\cite{song2020spatial} leans spatial and temporal dependencies simultaneously by a synchronous adjacency matrix and \textbf{GMSDR}~\cite{liu2022msdr} enhances recurrent models by explicitly using the features of different historical steps as the input.
    %\textbf{STGNCDE}~\cite{choi2022graph} and  \textbf{STGODE}~\cite{fang2021spatial} utilize neural control or ordinary differential equations to capture spatial and temporal dependencies.
    \textbf{MC Dropout}~\cite{wu2021quantifying} utilizes Monte Carlo dropout to estimate uncertainties. \textbf{TimeGrad}~\cite{rasul2021autoregressive} is an early-stage RNN-based diffusion method. \textbf{CSDI}~\cite{tashiro2021csdi}, \textbf{PriSTI}~\cite{liu2023pristi}, and \textbf{DiffSTG}~\cite{wen2023diffstg} leverages Transformers or U-Nets as the denoising networks. 
    \item \emph{Kriging}: \textbf{ADAIN}~\cite{cheng2018neural} is a sequential method using RNNs for temporal dependencies while \textbf{KCN}~\cite{appleby2020kriging} is a spatial model based on GNNs. \textbf{IGNNK}~\cite{wu2021inductive} leverage GNNs to spatial and temporal correlations simultaneously when predicting signals. \textbf{INCREASE}~\cite{zheng2023increase} explicitly divides spatial relations into several types and learns them separately. Note that diffusion models \textbf{CSDI}, \textbf{PriSTI}, and \textbf{DiffSTG} are also applicable to the kriging task.
\end{itemize}
Among the above baselines, MC Dropout, CSDI, DiffSTG, PriSTI, and TimeGrad are probabilistic models while the rest are deterministic networks. As shown in Fig.~\ref{fig:comp}, we further provide a comparison between USTD and other diffusion baselines. TimeGrad adopts individual diffusion processes to denoise targets at every time step, which incurs high computational costs. CSDI and DiffSTG directly feed conditions and denoised targets into the denoising network, making it hard to capture deterministic conditional relations. PriSTI concatenates conditions and targets, utilizing STGNNs to capture dependencies between them. However, the simple concatenation neglects the discrepancy between the two features, and STGNNs are trained together with denoising networks, which increases optimization difficulties. Our USTD solves these problems by only capturing conditional dependencies with a pre-trained STGNN encoder.
Please refer to Appendix~\ref{app:baseline} for baseline details.

\subsection{Experiment Setups and Evaluation Metrics}
For the forecasting task, we follow the settings~\cite{yu2018stgcn}, where the previous $T=12$ time steps are used to predict the subsequent $T^\prime=12$ steps. We partition all datasets into three temporal segments: training, validation, and testing with the split ratios of $6:2:2$ for PEMS-03~\cite{song2020spatial} and $7:2:1$ for the remaining datasets.
Regarding the kriging problem, we spatially partition the data into observed locations and unobserved locations at a ratio of $N:M=2:1$. The time window $T$ is also set to $12$.
For hyperparameters, the encoder consists of 6 layers mapping data into 64-size latent space, and the decoder has 3 layers. The graph sampling rate is fixed to 80\%.
TGA and SGA, trained individually, each has 2 layers with a 96-size channel, incorporating temporal~\cite{zhou2021informer} and spatial~\cite{dwivedi2020generalization} embeddings.
Following \cite{liu2023spatio}, we finetune the encoder when training the denoising modules. Diffusion settings are consistent with~\cite{tashiro2021csdi}. We conduct each experiment three times, reporting average results.
We utilize Mean Absolute Error (MAE) and Root Mean Squared Error (RMSE) as evaluation metrics. Moreover, for probabilistic methods, we report we report the Continuous Ranked Probability Score (CRPS) by sampling 8 predictions. CRPS utilizes target signals to measure the compatibility of an estimated probability distribution of the models. Small CRPS indicates better distribution estimates. Appendix~\ref{app:setup} provides more setup details. %The source code is released at \url{https://github.com/hjf1997/USTD}.

\newcolumntype{g}{>{\columncolor{white}}c}
\begin{table*}[h]
\centering
\caption{Results comparison on forecasting. The results are averaged over all time steps, where smaller metrics mean better performance. The bold is the overall best result. The underline indicates the second-best deterministic or probabilistic methods.}
\vspace{-1.em}
  \scalebox{1.02}{
  \centering
  \small
  \begin{tabular}[width=1.\linewidth]{lc||ccc|ccc|ccc|ccc}
  \shline
      \multirow{2}*{Method} & Dataset & \multicolumn{3}{c|}{PEMS-03} & \multicolumn{3}{c|}{PEMS-BAY} & \multicolumn{3}{c|}{AIR-BJ} & \multicolumn{3}{c}{AIR-GZ}\\
      \cline{3-14}
      & Metric & MAE & RMSE & CRPS & MAE & RMSE & CRPS & MAE & RMSE & CRPS & MAE & RMSE & CRPS \\
      \hline
      \hline
      STGCN & \multirow{7}*{\makecell{Deterministic\\ Method}} & 17.49 & 30.12 & \_ & 1.88 & 4.30 & \_ & 31.09 & 49.16 & \_ & 12.80 & 18.26 & \_ \\
      DCRNN & & 18.03 & 30.01 & \_ & 1.68 & 3.83 & \_ & 30.98 & 49.05 & \_ & 9.98 & 15.20 & \_ \\
      GWN & & 16.21 & 27.66 & \_ & \underline{1.67} & \underline{3.60} & \_ & 30.60 & {48.46} & \_ & \underline{9.90} & \underline{15.31} & \_ \\
      STSGCN & & 17.48 & 29.21 & \_ & 1.79 & 3.91 & \_ & 31.00 & 49.46 & \_ & 12.02 & 18.59 & \_ \\
      STGODE &  & 16.50 & 27.84 & \_ & 1.77 & \textbf{3.33} & \_ & 30.28 & \underline{48.26} & \_ & 10.89 & 15.73 & \_ \\
      STGNCDE &  & \underline{15.57} & 27.09 & \_ & {1.68} & 3.66 & \_ & \underline{30.45} & 49.17 & \_ & {10.10} & 15.90 & \_ \\
      GMSDR &  & 15.78 & \underline{26.82} & \_ & 1.69 & 3.80 & \_ & 32.15 & 51.08 & \_ & 11.78 & 17.72 & \_ \\
      \hline
      Improvement &  & 1.6\% & 2.8\% & \_ & 2.4\% & \_ & \_ & 1.2\% & 1.3\% & 4.1\% & 2.0\% & 1.5\% & \_ \\
      \hline
      \hline
      TimeGrad  & \multirow{6}*{\makecell{Probabilistic\\ Method}} & 21.55 & 36.57 & 0.101 & 2.62 & 5.30 & 0.034 & \underline{33.40} & 54.93 & \underline{0.363} & 15.45 & 21.93 & 0.376 \\
      MC Dropout & & 18.87 & 29.81 & 0.093 & 3.50 & 5.43 & 0.040 & 37.92 & 55.49 & 0.391 & 13.10 & 19.26 & \underline{0.290} \\
      CSDI & & 23.46 & 39.60 & 0.098 & 2.67 & \underline{4.10} & 0.031 & 38.94 & 57.81 & 0.417 & 14.78 & 22.24 & 0.361 \\
      DiffSTG & & \underline{17.58} & \underline{28.75} & 0.095 & \underline{2.03} & 4.22 & \underline{0.025} & 38.03 & 56.87 & 0.373 & \underline{13.06} & \underline{18.25} & 0.319  \\
      PriSTI & & 22.30 & 37.58 & \underline{0.092} & 2.51 & 3.99 & 0.026 & 36.81 & \underline{54.34} & 0.388 & 14.04 & 21.03 & 0.352 \\
      \rowcolor{Gray}
      USTD & & \textbf{15.32} & \textbf{26.06} & \textbf{0.087} & \textbf{1.63} & {3.55} & \textbf{0.022} & \textbf{30.09} & \textbf{47.65} & \textbf{0.348} & \textbf{9.70} & \textbf{15.08} & \textbf{0.257} \\    
    \hline
    Improvement & & 12.8\% & 9.3\% & 5.4\% & 19.7\% & 13.4\% & 12.0\% & 10.0\% & 12.3\% & 4.1\% & 25.7\% & 17.4\% & 11.4\% \\
  \shline
\end{tabular}
}
\label{tab:forecasting}
\end{table*}

\begin{table*}[h]
\centering
\caption{Performance comparison of USTD and baselines on the kriging task.}
\vspace{-1.em}
  \scalebox{1.02}{
  \centering
  \small
  % \begin{threeparttable}
  \begin{tabular}[width=1.\linewidth]{lc||ccc|ccc|ccc|ccc}
  \shline
      \multirow{2}*{Method} & Dataset & \multicolumn{3}{c|}{PEMS-03} & \multicolumn{3}{c|}{PEMS-BAY} & \multicolumn{3}{c|}{AIR-BJ} & \multicolumn{3}{c}{AIR-GZ}\\
      \cline{3-14}  %\cline{5-7}
      & Metric & MAE & RMSE & CRPS & MAE & RMSE & CRPS & MAE & RMSE & CRPS & MAE & RMSE & CRPS \\
      \hline
      \hline
      ADAIN & \multirow{4}*{\makecell{Deterministic\\ Method}} & 16.93 & 38.47 & \_ & 3.35 & 6.32 & \_ & 15.04 & 29.59 & \_ & 10.28 & 15.37 & \_  \\  
      KCN  & & 16.01 & 37.35 & \_ & 2.63 & 5.17 & \_ & 14.88 & 28.98 & \_ & 9.81 & 15.06 & \_  \\  
      IGNNK &  & 15.50 & 36.17 & \_ & 2.30 & 4.58 & \_ & \underline{13.86} & \underline{27.48} & \_ & 8.95 & 13.76 & \_  \\  
      INCREASE  & & \underline{15.34} & \underline{35.95} & \_ & \underline{2.19} & \underline{4.51} & \_ & 14.10 & 28.48 & \_ & \underline{8.82} & \underline{13.58} & \_  \\
      \hline
      Improvement & & 4.0\% & 2.8\% & \_ & 10.5\% & 6.4\% & \_ & 4.0\% & 1.4\% & \_ & 2.4\% & 3.5\% & \_ \\
      \hline
      \hline
      CSDI & \multirow{4}*{\makecell{Probabilistic\\ Method}} & 15.09 & 35.73 & 0.082 & 2.35 & 4.62 & 0.032 & 14.14 & 28.97 & 0.135 & 9.62 & 14.65 & 0.246  \\  
      DiffSTG & & \underline{14.90} & 35.89 & 0.079 & 2.20 & 4.35 & 0.028 & 14.01 & \underline{27.61} & 0.138 & 9.91 & 15.13 & 0.257\\
      PriSTI &  & {14.95} & \underline{35.42} & \underline{0.076} & \underline{2.06} & \underline{4.29} & \underline{0.026} & \underline{13.78} & 27.86 & \underline{0.133} & \underline{8.98} & \underline{14.13} & \underline{0.231}  \\  
      \rowcolor{Gray}
      USTD  & & \textbf{14.73} & \textbf{34.94} & \textbf{0.071} & \textbf{1.96} & \textbf{4.22} & \textbf{0.025} & \textbf{13.30} & \textbf{27.09} & \textbf{0.130} & \textbf{8.61} & \textbf{13.10} & \textbf{0.213}  \\  
      \hline
      Improvement & & 1.1\% & 1.4\% & 6.6\% & 4.9\% & 1.6\% & 3.8\% & 3.5\% & 1.8\% & 2.3\% & 4.1\% & 7.2\% & 7.8\% \\
  \shline
\end{tabular}
% \end{threeparttable}
}
\label{tab:kriging}
\end{table*}

\subsection{Performance Comparison}
Table \ref{tab:forecasting} and \ref{tab:kriging} compare the performances of USTD and the baselines on the two tasks. From the tables, we observe that USTD consistently outperforms all probabilistic methods. Compared to the second-best probabilistic methods, our model decreases CRPS by up to 12.0\% on forecasting and 7.8\% on kriging, which indicates our model has a better ability to capture uncertainties. 
When it comes to deterministic baselines, our model still can surpass them only except for the RMSE of STGODE on PEMS-BAY. Numerically, compared to the second-best models, USTD reduces the MAE by up to 2.4\% and 10.5\% on two tasks, respectively. As previous approaches found probabilistic methods are hard to surpass their deterministic counterparts~\cite{wen2023diffstg}, the performances of USTD are compelling.

From the tables, we also have the following observations and explanations: (i) Probabilistic baselines struggle to outperform deterministic networks, especially for forecasting that is more well-explored, which is consistent with the previous findings. As these baselines optimize the encoder together with denoising networks, the models are required to learn both deterministic and probabilistic dependencies simultaneously. This increases training difficulties and hampers their performance.
(ii) Probabilistic models gain improved performance on the spatial prediction task--kriging, particularly for PriSTI. This is likely because the graph information it utilized, defining correlations between conditional and target nodes, serves as an extra prior that facilitates spatial learning.
(iii) Our USTD, as a probabilistic method, outperforms the deterministic baselines. This suggests the pre-trained encoder learns representative conditional dependencies, which alleviates the learning difficulties for the subsequent denoising module. 

\subsection{Inference Time}
DDPM's one-stop sampling manner of $q(x_k|x_0)$ during training casts off the need for recurrent sampling, which ensures its training efficiency. However, it still bears a notorious long inference time during testing, partially due to recurrent sampling on the chain. As our encoder squashes the temporal dimension with TCNs, the denoising modules take in the resultant low-dimensional representations, resulting in less computation demand and a faster inference time. To demonstrate this, we compare the inference time of USTD with CSDI and PriSTI, where the time indicates the sampling cost of one prediction, and conduct the study on two datasets with the largest number of nodes.
From Table~\ref{tab:computation}, we find USTD enjoys a faster computation speed, which reduces the time by at most 47.3\% compared to the second best. As our model follows the same sampling procedure with the baselines but relies on the compressed conditional information, this finding suggests that a more efficient denoising network can significantly reduce the inference time.  

\begin{table}[h]
\caption{Inference time (seconds) of diffusion models.}
\vspace{-1.em}
\centering
  \scalebox{1.1}{
  \centering
  \small
  \begin{tabular}[width=1.\linewidth]{l||cc|cc}
  % \shline
    \shline
      \multirow{2}*{Method} & \multicolumn{2}{c|}{PEMS-03} & \multicolumn{2}{c}{PEMS-BAY}\\
      \cline{2-5} 
      & Forecasting & Kriging & Forecasting & Kriging \\      
      \hline
      \hline
      CSDI  & 0.877 & 0.898 & 0.931 & 0.852 \\  
      PriSTI  & 1.133 & 1.045   & 1.047 & 1.022   \\  
      \rowcolor{Gray}
      USTD  & 0.501 & 0.496 & 0.490 & 0.487  \\  
      \hline
      Improvement & 43.5\% & 44.8\% & 47.3\% & 42.8\% \\
  \shline
\end{tabular}
}
\label{tab:computation}
\end{table}

\subsection{Case Study}
\begin{figure}
\setlength{\abovecaptionskip}{-0.01mm}
\centering
\subfigcapskip=-5pt
\subfigure[Node 330 in PEMS-03]{
\begin{minipage}[t]{0.5\linewidth}
\centering
\includegraphics[width=1.0\linewidth]{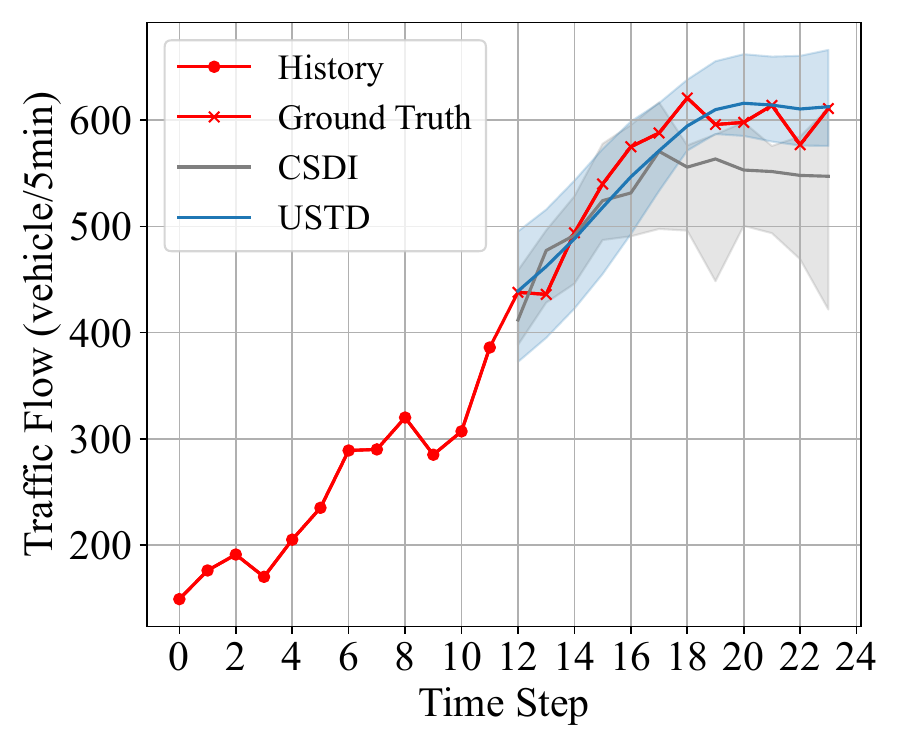}
%\caption{fig1}
\end{minipage}%
}%\hskip -8pt
\subfigure[Node 324 in PEMS-BAY]{
\begin{minipage}[t]{0.5\linewidth}
\centering
\includegraphics[width=1.0\linewidth]{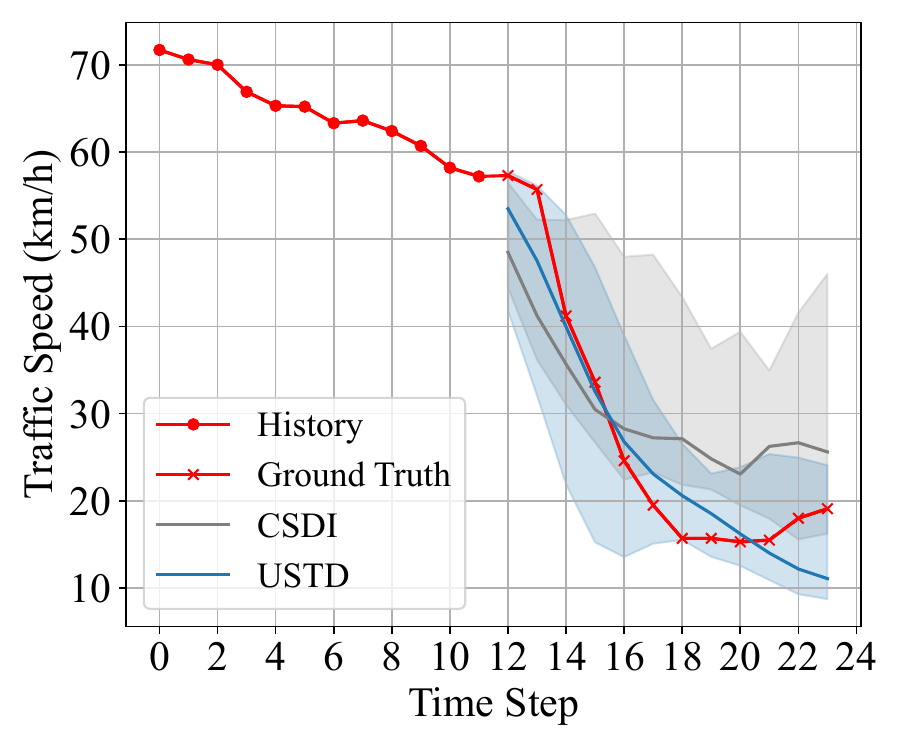}
% (b) Size of hidden states for ASGGRU.
%\caption{fig2}
\end{minipage}
}
\vskip -5pt
\subfigure[Node 27 in AIR-BJ]{
\begin{minipage}[t]{1\linewidth}
\centering
\includegraphics[width=1.0\linewidth]{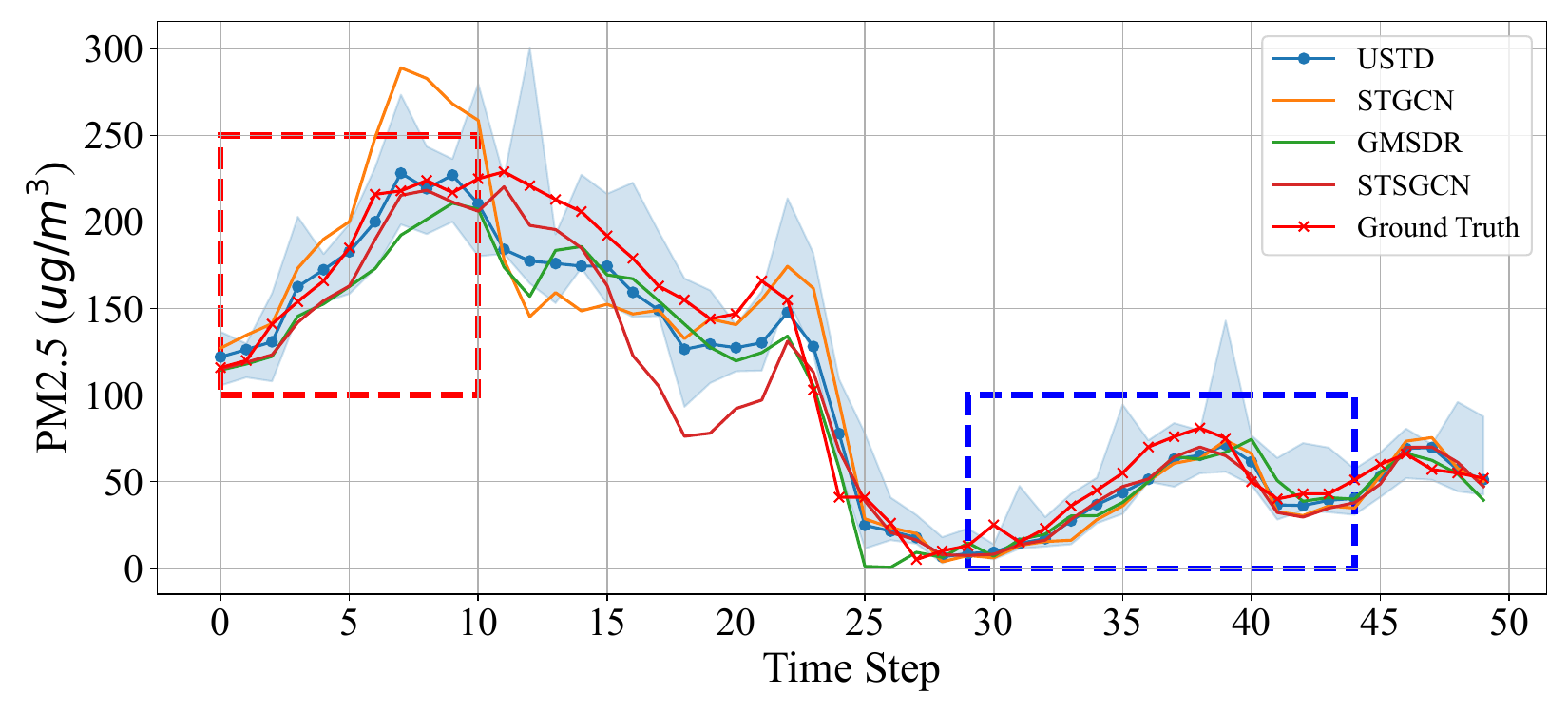}
% (c) Number of input sequence length $T$.
%\caption{fig2}
\end{minipage}
}%\hskip -10pt
\centering
\caption{Forecasting visualizations on three datasets.}
% (a) and (b) are predictions of 12 successive time steps based on historical signals. (c) is the combination of 1-step ahead predictions.
\label{fig:case}
\end{figure}

To gain insights into the uncertainty estimates and prediction accuracy, we conduct case studies to visualize the results of USTD and the baselines on the forecasting task. Fig.~\ref{fig:case}(a) and \ref{fig:case}(b) present the prediction of $T^\prime=12$ future time steps based on $T=12$ historical steps, from which we find USTD numerically outperforms the diffusion model CSDI, as our predictions are closer to the ground truth. In addition, regarding uncertainty estimates, our model demonstrates higher reliability because the shadow straps encompass the true signals with a narrower sampling range. This means that USTD is able to extract high-quality spatio-temporal dependencies from historical data, probably benefiting from the pre-trained encoder. In Fig.~\ref{fig:case}(c) we illustrate the successive prediction values of 1-step ahead forecasting and compare our model with deterministic baselines. As the red block indicates, USTD enjoys superiority over the other baselines. In addition, in cases where USTD performs similarly to the baselines, as shown in the blue block, its probabilistic nature provides additional information regarding the uncertainty of predictions indicated by the shadow that covers the ground truth. This uncertainty estimate facilitates decision-making in real-world trustworthy applications. 

\subsection{Abaltion Study}
\begin{figure}[h]
  \centering
  \includegraphics[width=1.03 \linewidth]{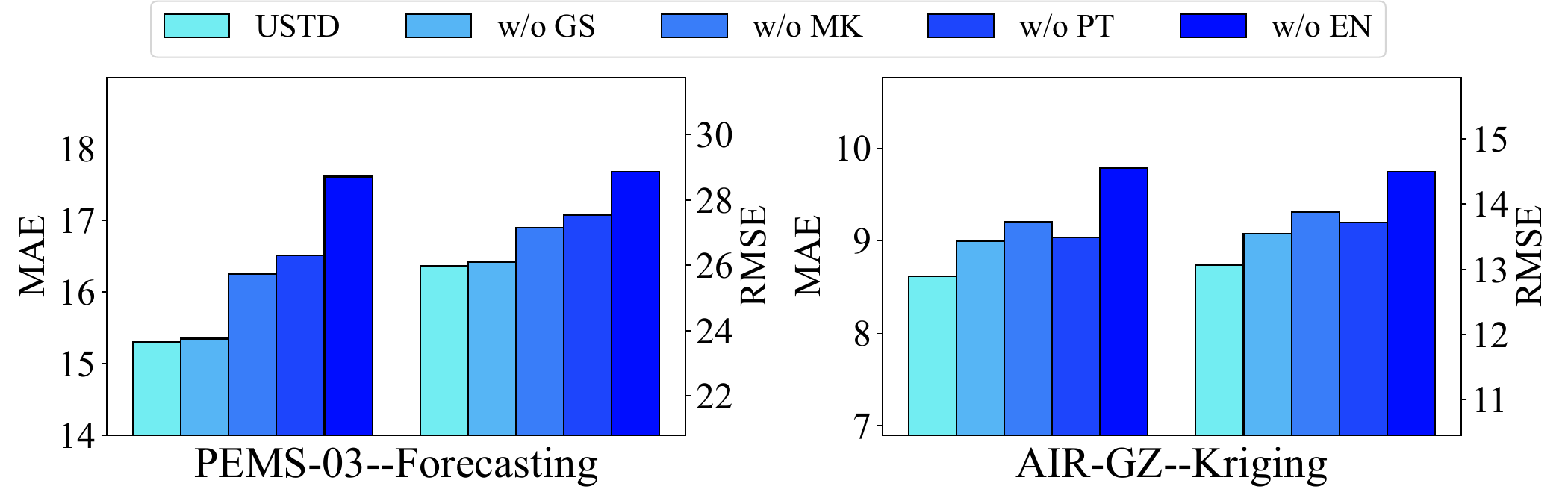}
  % \vspace{-1.5em}
  \caption{Effects of the pre-trained encoder.}
    \label{fig:ablation1}
    % \vspace{-0.5em}
\end{figure}

We evaluate the effects of each component in USTD. We only adjust the specific settings while keeping the others unchanged. %To evaluate the effects under different data and tasks, 
To comprehensively evaluate the model under different datasets and tasks, we use PEMS-03 for forecasting and AIR-GZ for kriging.
\paragraph{Effects of Pre-Trained Encoder}
To study the effects of the influence of the pre-training strategy, graph sampling, and masking on the encoder, we adopt the following variants: (a) w/o EN: we discard the encoder and the pre-training stage. Only the denoising network is employed. (b) w/o PT: we train the encoder and the denoising network together in an end-to-end way. (c) w/o MK: We pre-train the encoder without the masking mechanism. (d) w/o GS: Graph sampling is not involved in the training. Fig.~\ref{fig:ablation1} shows the results on the two tasks. Removing the encoder leads to the worst performance deterioration, verifying the importance of effectively modeling conditional dependencies. Then, pre-training the encoder also plays a crucial role, which pre-extracts dependencies and reduces learning difficulties for the denoising module.
Moreover, masking benefits both forecasting and kriging tasks and prevents the model from obtaining trivial solutions. Then, graph sampling boosts the model's performance in kriging, as the encoder is required to adapt to varied graph structures. 

\begin{figure}[h]
  \centering
  \includegraphics[width=1.03 \linewidth]{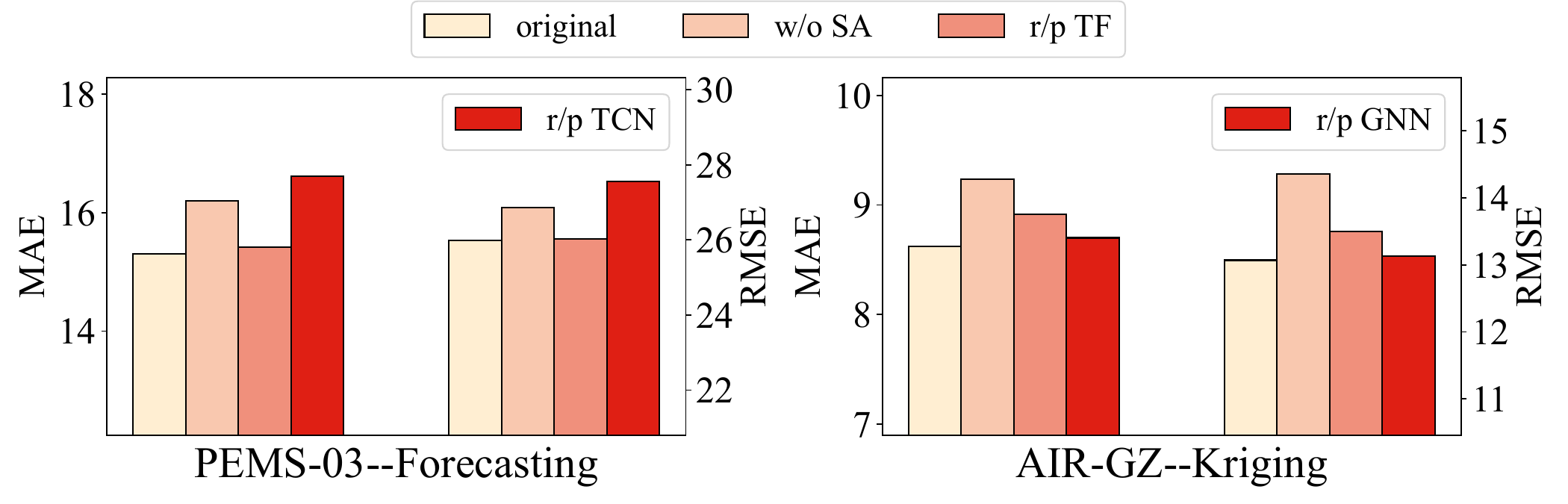}
  % \vspace{-1.5em}
  \caption{Effects of the gated attention networks.}
    \label{fig:ablation2}
\end{figure}

\paragraph{Effects of Gated Attention Networks}
We consider four variants to examine the efficacy of gated attention networks. (a) r/p TCN: We replace the cross-attention in TGA with a temporal convolutional network. (b) r/p GNN: SGA's cross-attention is substituted with a GNN. (c) w/o SA: The self-attention block is detached. (d) r/p TF: This variant replaces the two attentions with a standard Transformer layer.
The results are shown in Fig.~\ref{fig:ablation2}, in which we find that r/p TCN and w/o SA lead to the largest performance degradation for forecasting and kriging, respectively. This is because TCN's kernel requires a uniform temporal evolution that is not achieved by concatenating learned conditional representations and the targets to be denoised. 
Moreover, self-attention plays an exclusive role in capturing correlations among the target nodes, which can be crucial for kriging.
Both r/p TF and r/p GNN cause a slight performance deterioration, demonstrating the effectiveness of the gated attention and the encoder’s ability to capture spatial relations.

\subsection{Hyperparameter Study}
\begin{figure}[h]
\setlength{\abovecaptionskip}{-0.01mm}
\centering
\subfigure[Hyperparameters of forecasting on PEMS-03]{
\includegraphics[width=1 \linewidth]{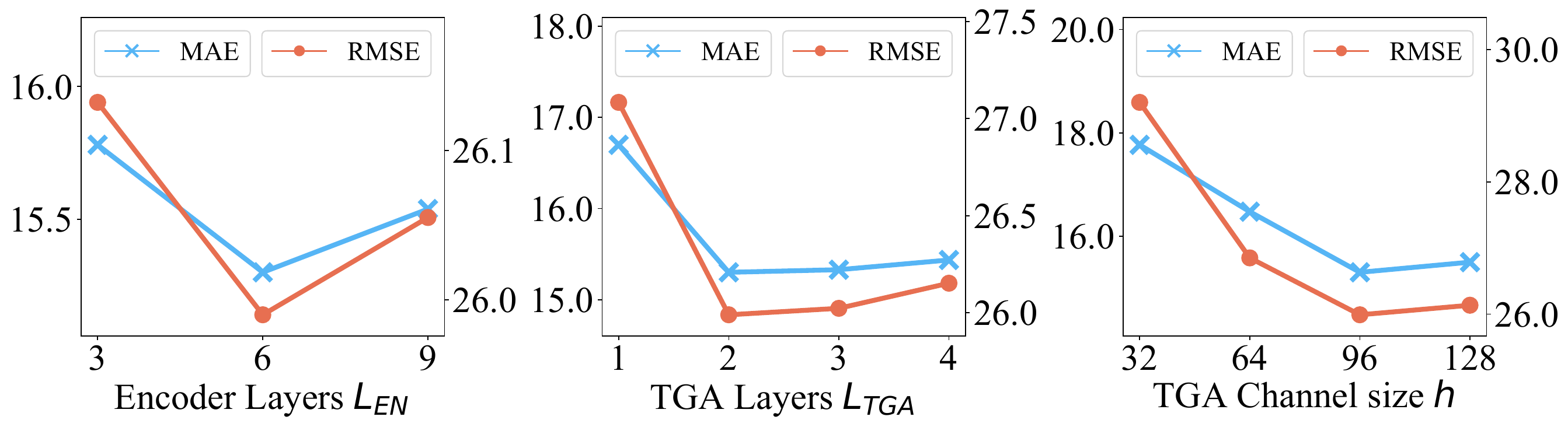}
}
\vskip -2pt

\subfigure[Hyperparameters of kriging on AIR-GZ]{
\includegraphics[width=1 \linewidth]{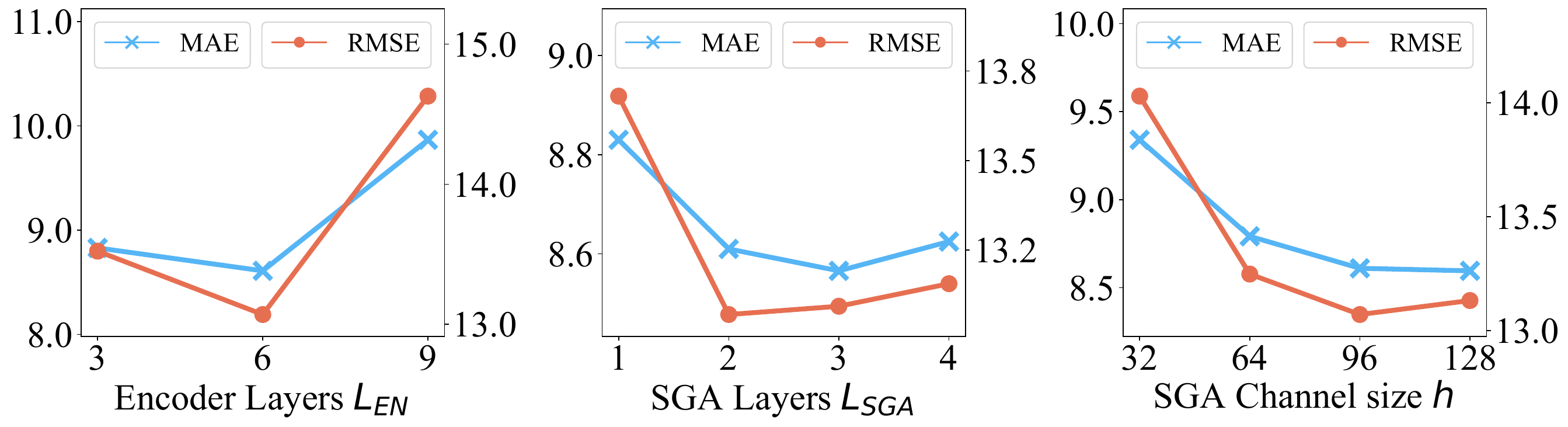}
}
\caption{Effects of USTD's hyperparameters.}
\label{fig:hyper}
\end{figure}

We first study the effects of the number of encoder layers $L_{EN}$, TGA blocks $L_{TGA}$, and the channel size $h$ of the blocks on forecasting. Fig.~\ref{fig:hyper}(a) reports the results on PEMS-03 and we have the following observations: 1) Stacking 6 encoder layers achieves the best, while the model is overfitting clearly with $L_{EN}=9$. 2) The performance of USTD is not sensitive when $L_{TGA}>1$ and $h>64$. Therefore, the TGA module has 2 layers with a channel size of 96. Then, we evaluate the effects on the kriging task using AIR-GZ and change the number of SGA layers and the channel size. The results are shown in Fig.~\ref{fig:hyper}(b), from which we find the model gets the best results when $L_{EN}=6$, $L_{TGA}=2$ and $h=96$, aligning with the hyperparameters of forecasting. The sensitivity of the encoder and denoising networks can be explained by its optimization method. Overfitting is a common phenomenon for deterministic networks like the encoder. On the contrary, as the denoising networks are trained through variational approximation, they have more opportunities to search in the training space to skip local minima. 

\section{Related Work}
\subsection{Spatio-Temporal Graph Neural Networks}
Spatio-temporal graph neural networks (STGNNs) are a popular paradigm for spatio-temporal data modeling nowadays. They serve as feature extractors to capture complex spatial and temporal dependencies in conditional information. Mainly following two categories, they first resort to graph neural networks to model spatial correlations, then combine with either temporal convolutional networks~\cite{wu2020connecting,wu2019graph} or recurrent neural networks~\cite{xu2019spatio,shu2020host} for temporal relationship capturing. The learned conditional patterns can be used for various downstream learning tasks. 
For instance, \cite{li2018dcrnn_traffic} proposed an encoder-decoder architecture for traffic flow prediction. \cite{deng2021graph} introduced a structure learning approach by GNNs for multivariate time-series anomaly detection. \cite{shu2020host} employed graph LSTMs to model multi-level dependencies for group activity recognition. %\cite{fang2021spatial,choi2022graph,liu2022msdr,song2020spatial,wu2019graph} all propose various mechanisms based on STGNNs to boost the capturing of dependencies. 
However, most STGNNs mainly focus on deterministic dependency modeling while neglecting the importance of uncertainty estimates. 

\subsection{Spatio-Temporal Graph Learning} 
Spatio-temporal graph learning involves various downstream tasks. In this work, we focus on two crucial problems: forecasting and kriging. The forecasting models rely on STGNNs to capture spatio-temporal dependencies and propose various methods to boost their learning effectiveness. For example, STSGCN~\cite{song2020spatial} captures spatial and temporal correlations simultaneously by a synchronous modeling mechanism. STGODE~\cite{fang2021spatial} and STGNCDE~\cite{choi2022graph} rely on the neural differential equations to model long-range spatial-temporal dependencies. GMSDR~\cite{liu2022msdr} introduces an RNN variant that explicitly maintains multiple historical information at each time step.
Similar to forecasting, kriging methods still count on STGNNs but emphasize the spatial relations between observed and target nodes, the only information available for target nodes~\cite{appleby2020kriging,wu2021inductive}. These relations are typically modeled by various graph aggregators.  
KCN~\cite{appleby2020kriging} utilizes a mask to indicate the target node in the aggregator. 
SATCN~\cite{wu2021spatial} extracts statistical properties of the target nodes while INCREASE~\cite{zheng2023increase} divides spatial relations into several types and captures them separately. It also utilizes external features like POIs as auxiliary information. Although these methods achieve promising performances on their tasks, their dedicated designs are tailored for individual problems, which limits their versatility on other problems. To fill the gap, we move towards unifying the tasks by proposing a shared encoder and task-specific decoders that are holistically designed. 

\subsection{Spatio-Temporal Diffusion Models} 
DDPMs are initially proposed for computer vision as a generative model~\cite{ho2020denoising,yang2022diffusion}. By decomposing a complex distribution into a combination of simple Gaussian distributions and generating samples by recovering the distribution step by step, DDPMs are empowered with strong capabilities in modeling complex data distributions.
Recently, the community has started to explore its potential for spatio-temporal tasks~\cite{yang2024survey}. TimeGrad~\cite{rasul2021autoregressive} is a pioneer diffusion work for forecasting which utilizes RNNs to model temporal dependencies. Due to 
its autoregressive nature, it suffers from a long inference time.
%which suffers from a huge inference time due to its autoregressive nature. 
DiffSTG~\cite{wen2023diffstg} utilizes a U-net architecture to speed up sampling.
CSDI~\cite{tashiro2021csdi} is proposed for data imputation based on Transformers and PriSTI~\cite{liu2023pristi} further adds GNNs to capture graph dependencies. SpecSTG~\cite{lin2024specstg} generates Fourier representations of future signals.
However, they neglect the importance of pre-extracting conditional patterns to benefit denoising networks, making it difficult for them to outperform deterministic models.

\section{Conclusion And Future Work}
We present USTD as the first step towards unifying spatio-temporal graph learning. The motivation lies in the insight that dependencies of conditional spatio-temporal data are universal and should be shared by different downstream tasks. Therefore, our model leverages a pre-trained encoder to effectively extract conditional dependencies for the subsequent denoising networks.
Then, taking in the learned representations, the holistically designed denoising modules TGA and SGA utilize attention blocks to generate predictions for forecasting and kriging, respectively.
Compared to both deterministic and probabilistic methods, our USTD achieves state-of-the-art performances on both tasks while providing valuable uncertainty estimates. 
We summarize two research directions for future endeavors. First, we can apply USTD to other spatio-temporal graph applications such as imputation or classification. Second, it is promising to explore the feasibility of training a single model that can be applied to various downstream tasks.

\begin{acks}
This research is supported by Singapore Ministry of Education Academic Research Fund Tier 2 under MOE's official grant number T2EP20221-0023. This paper is also supported by the National Natural Science Foundation of China (No. 62402414).
\end{acks}

\bibliographystyle{ACM-Reference-Format}
\balance
\bibliography{sample-base}

\clearpage
\appendix
\section{Mathematical Notation}
\label{app:notation}
% We define important notations used in the paper in Table~\ref{tab:notation}.
\begin{table}[h]
\caption{Mathematical notations. Dims indicate dimension.}
\vspace{-1.em}
\scalebox{0.95}{
  \begin{tabular}[width=0.75\linewidth]{lll}
    \shline
      Notation & Dims & Description \\
      \hline
      \hline
    $T$ & $1$ & Historical time window\\
    $T^\prime$ & $1$ & Future time window\\
    $N$, $M$ & $1$, $1$ & Observed/Unobserved nodes\\
    $d_x$, $d_y$, $d_h$ & $1$ & Feature dimension\\
    $t$, $i$ & $1$, $1$ & Time, node index\\
    $\tau$ & $1$  & length of $H$\\
    $k$ & $1$ & Index of diffusion step\\
    \hline
    $x^t_i$ & $d_x$ & Signals of node $i$ at time $t$\\
    \hline
    $X_i$ & $T\times d_x$ & Conditional data of node $i$\\
    $X^t$ & $N\times d_x$ & Nodes signals at time $t$\\
    $Y^{T+1:T+T^\prime}$ & $N\times T^\prime \times d_y$ & Forecasting target\\
    $Y_{N+1:N+M}$ & $M\times T \times d_y$ & Kriging target\\
    $H$ & $N\times \tau \times d_h$ & Conditional representations\\
  \shline
\end{tabular}
}
\label{tab:notation}
\end{table}

\section{Training and Inference Details}
\label{app:training}
% We provide training and inference details of the denoising modules.

\paragraph{Training} The primary goal of the probabilistic model to is maximize its likelihood $\log p_\theta(Y|H, \mathcal{G})$, where $Y$ is the target and $H=\operatorname{Enc}_{\phi}(X, \mathcal{G})$ is the conditional representation learned by the pre-trained encoder $\operatorname{Enc}_{\phi}(\cdot)$. The likelihood can be optimized by randomly sampling conditional information and the target from the dataset. However, this goal cannot be achieved trivially due to the nonlinearity of the model. As a makeshift, we train the model through variational approximation by minimizing its negative evidence lower bound (ELBO). Formally, we first obtain representation $H$ from the encoder. Then, the sampled diffusion step $k$ and noise $\epsilon$ are utilized to calculate the corrupted target $Y_k=\sqrt{\alpha_k} Y_0+\sqrt{1-\alpha_k} \epsilon$. Lastly, the denoising module tasks in $Y_k$, $k$, conditional representations $H$, and the graph structure $\mathcal{G}$ to estimate the noise $\epsilon$. 

\paragraph{Inference} After the model is trained, we aim to predict target variables based on conditional representations. Given the condition, we first adopt the encoder to learn its representations $H$. Then, starting from a Gaussian noise $Y_K\sim\mathcal{N}(0, \boldsymbol{I})$, the denoising network recovers the target $Y_0$ gradually, following the reverse Markov chain. The process can be repeated many times, with each sample representing one possible result. The range of all generated samples suggests uncertainty estimates of the target.

\section{USTD Architectures}
\label{app:ustd}
\begin{figure}[h]
  \centering
  \includegraphics[width=0.82 \linewidth]{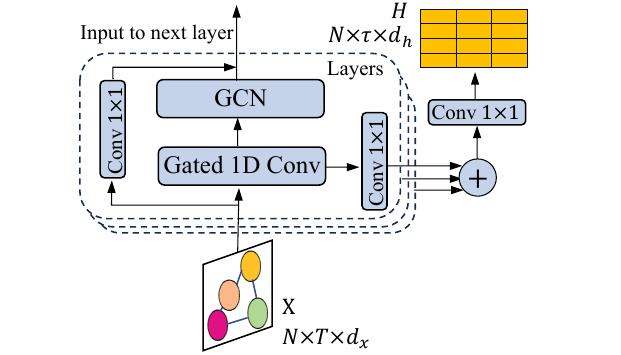}
  \caption{The pre-trained encoder contains spatio-temporal layers, with gated 1D convolutions for temporal capturing and graph convolution networks for spatial modeling.}
    \label{fig:encoder_arch}
\end{figure}

\paragraph{Pre-Trained Encoder}
The encoder is a combination of spatio-temporal layers~\cite{wu2019graph}, as shown in Fig.~\ref{fig:encoder_arch}. To be specific, in each layer, we use a gate 1D convolution to model temporal dependencies. Then, a graph convolutional network is employed to capture spatial relations given the graph structure. We add a residual connection in the layer through a $1\times 1$ convolution. After the bottom layer, outputs of all layers are aggregated by a skip connection with a $1\times 1$ convolution. Finally, a linear layer is used to get conditional representations denoted by $H$.

\begin{figure}[h]
  \centering
  \includegraphics[width=0.8 \linewidth]{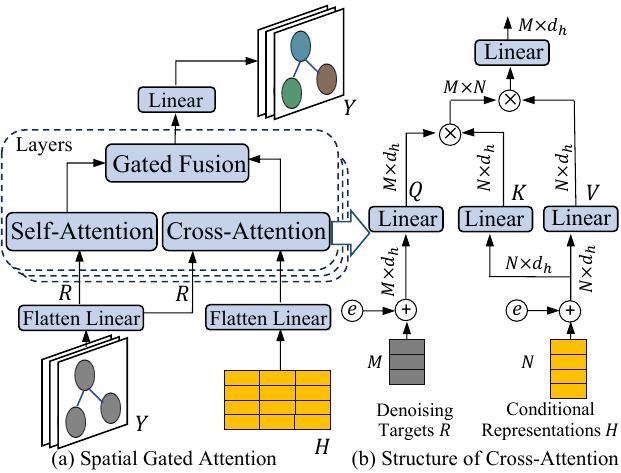}
  \caption{(a) Spatial Gated Attention. (b) Pipeline of the cross-attention, where $e$ denotes space and diffusion embeddings.}
    \label{fig:sga}
\end{figure}

\paragraph{Spatial Gated Attention Network}
As shown in Fig.~\ref{fig:sga}, the cross-attention is conducted only on the spatial dimension that is more important for kriging. Specifically, we use linear layers to get the flattened embeddings of the previously predicted results $Y$ and learned representations $H$, denoted by $R\in\mathbb{R}^{M\times d_h}$ and $H\in\mathbb{H}^{N\times d_h}$. Then, cross-attentions are employed to capture dependencies between target nodes and representations, while self-attentions are used to capture correlations within the target nodes. Next, a fusion gate aggregates features, and a linear layer is used to estimate the noises of the current diffusion step. We add space and diffusion embeddings to the network, providing graph and diffusion information. 

\begin{figure*}
\setlength{\abovecaptionskip}{-0.01mm}
\setlength{\belowcaptionskip}{-10pt}
\centering
\subfigcapskip=-2pt
\subfigure{
\begin{minipage}[t]{0.22\linewidth}
\centering
\includegraphics[width=1.0\linewidth]{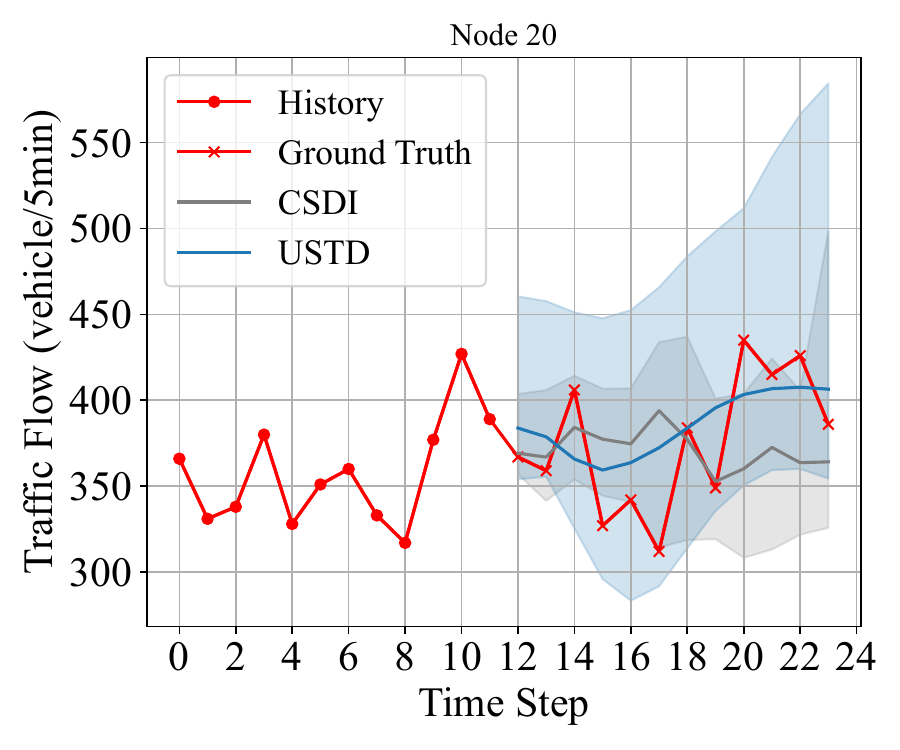}
%\caption{fig1}
\end{minipage}
}%\hskip -4mm
\subfigcapskip=-2pt
\subfigure{
\begin{minipage}[t]{0.22\linewidth}
\centering
\includegraphics[width=1.0\linewidth]{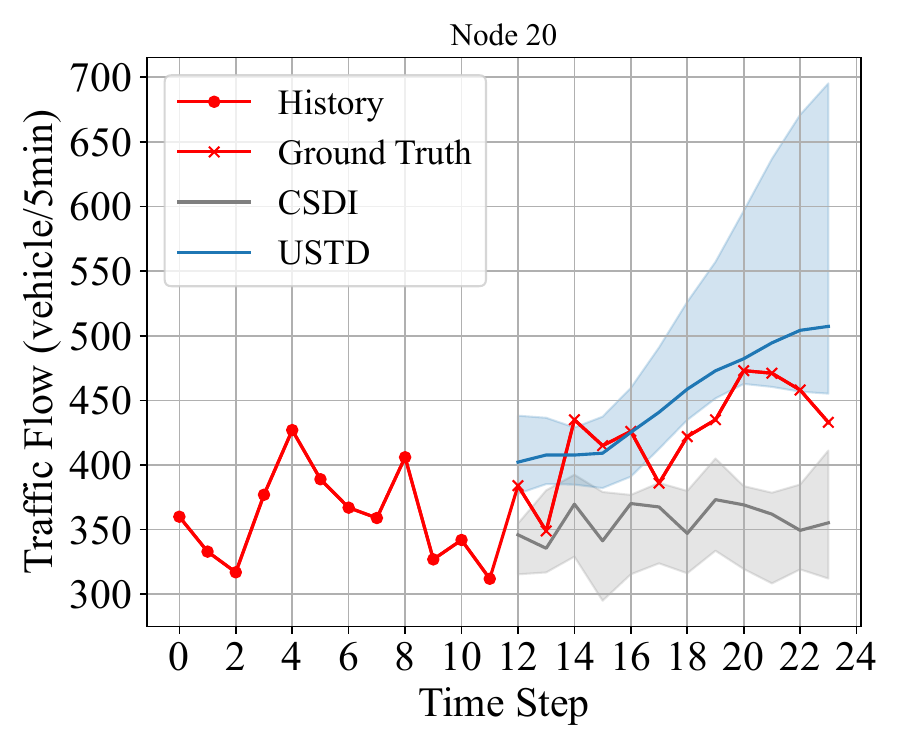}
%\caption{fig1}
\end{minipage}
}%\hskip -4mm
\subfigcapskip=-2pt
\subfigure{
\begin{minipage}[t]{0.22\linewidth}
\centering
\includegraphics[width=1.0\linewidth]{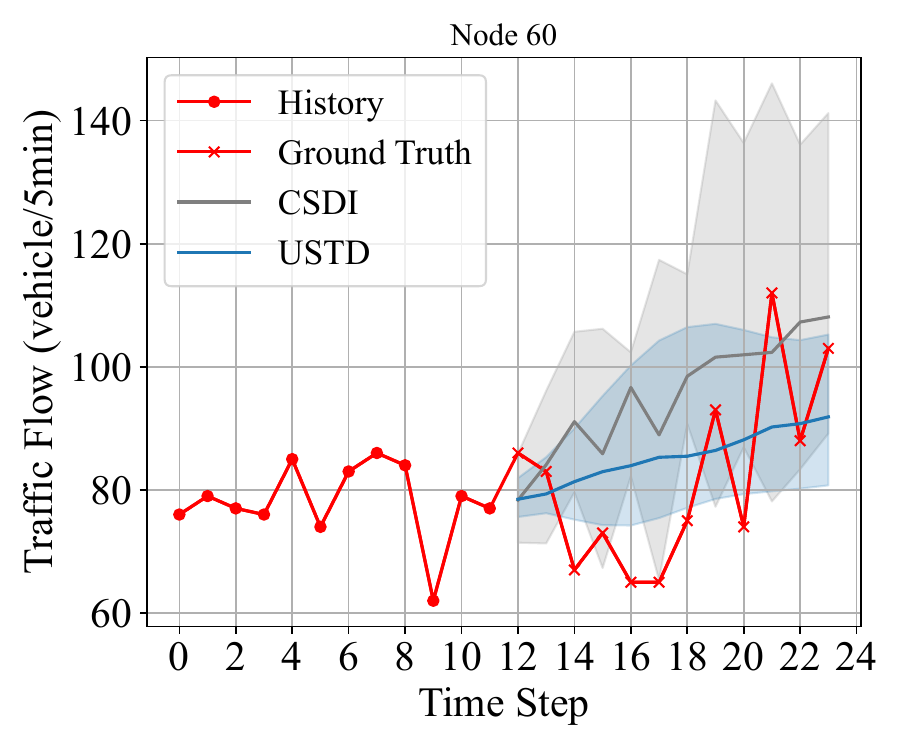}
%\caption{fig1}
\end{minipage}
}%\hskip -4mm
\subfigcapskip=-2pt
\subfigure{
\begin{minipage}[t]{0.22\linewidth}
\centering
\includegraphics[width=1.0\linewidth]{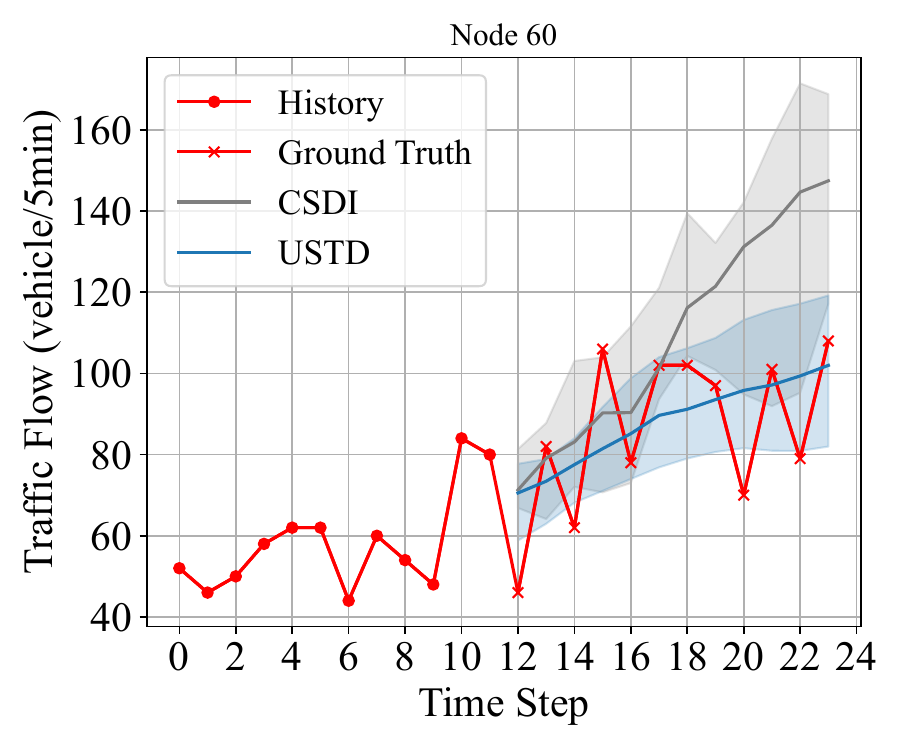}
\end{minipage}%
}

\caption{Comparison of forecasting results between USTD and CSDI on the PEMS-03 dataset.}
\label{fig:vis03}
\end{figure*}
%%%%%%%%%%%%%%%%%%%%%%%%%%%%%%%%%%%%%%%%%%%%%%%%%%%%%%
%%%%%%%%%%%%%%%%%%%%%%%%%%%%%%%%%%%%%%%%%%%%%%%%%%%%%%%
\begin{figure*}
\setlength{\abovecaptionskip}{-0.01mm}
\centering
\subfigcapskip=-2pt
\subfigure{
\begin{minipage}[t]{0.22\linewidth}
\centering
\includegraphics[width=1.0\linewidth]{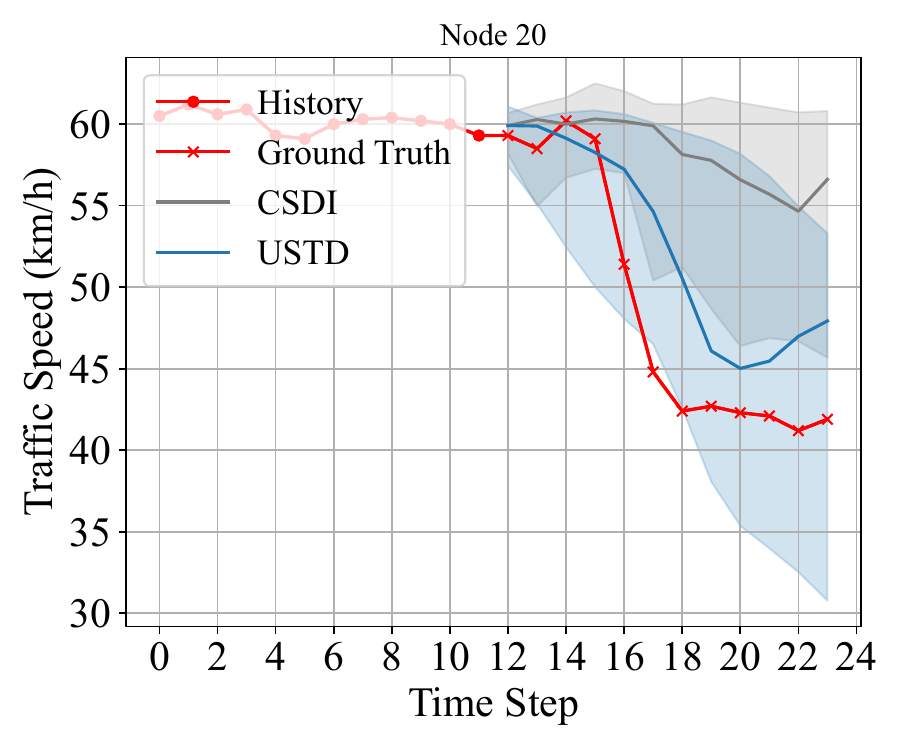}
%\caption{fig1}
\end{minipage}
}%\hskip -4mm
\subfigcapskip=-2pt
\subfigure{
\begin{minipage}[t]{0.22\linewidth}
\centering
\includegraphics[width=1.0\linewidth]{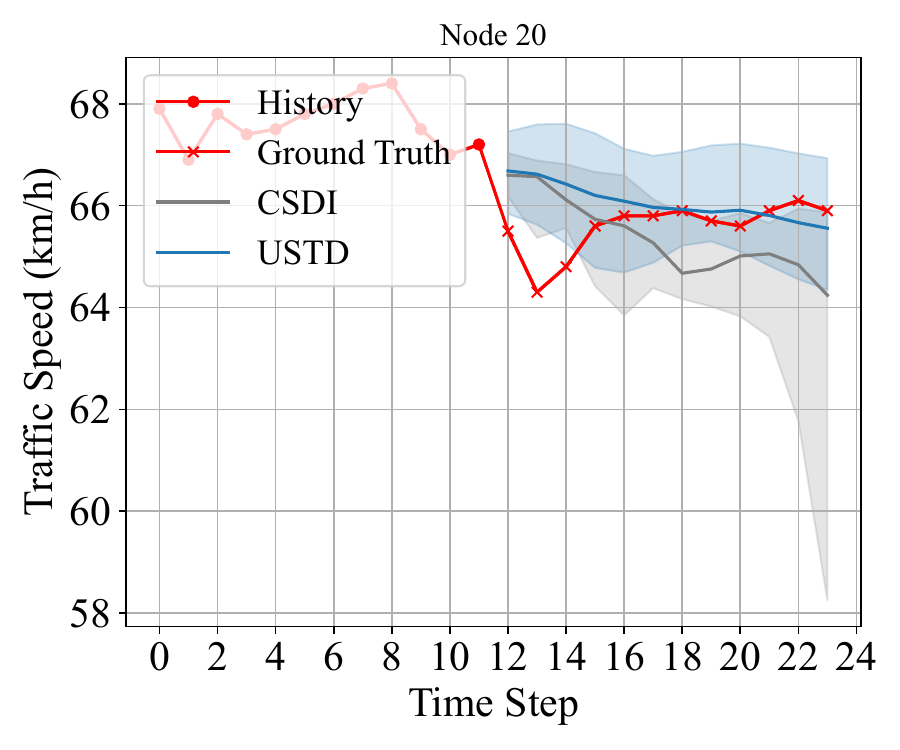}
%\caption{fig1}
\end{minipage}
}%\hskip -4mm
\subfigcapskip=-2pt
\subfigure{
\begin{minipage}[t]{0.22\linewidth}
\centering
\includegraphics[width=1.0\linewidth]{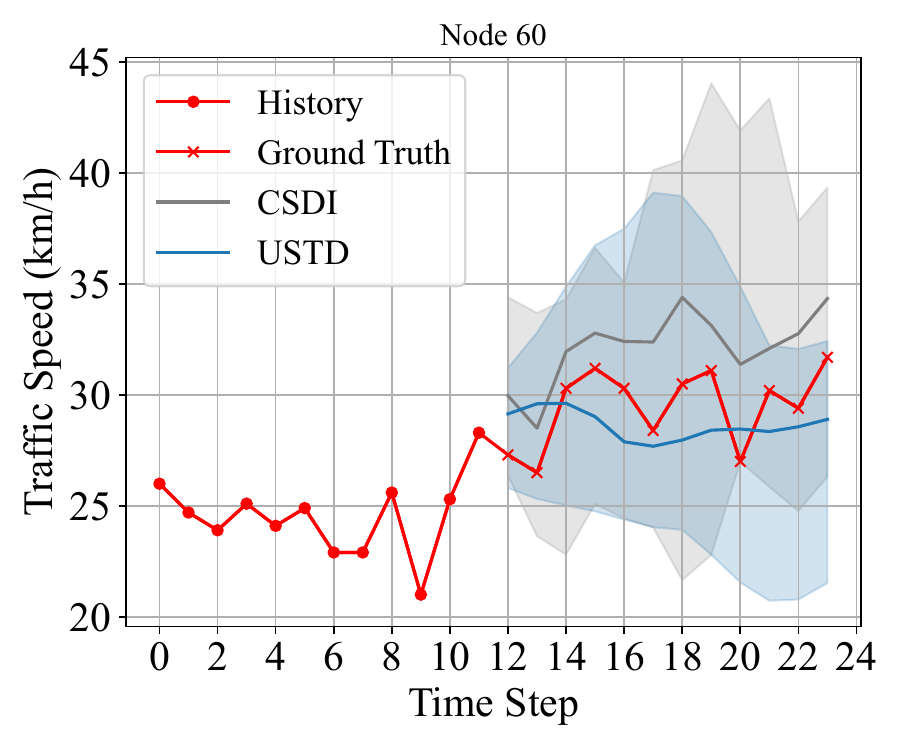}
%\caption{fig1}
\end{minipage}
}%\hskip -4mm
\subfigcapskip=-2pt
\subfigure{
\begin{minipage}[t]{0.22\linewidth}
\centering
\includegraphics[width=1.0\linewidth]{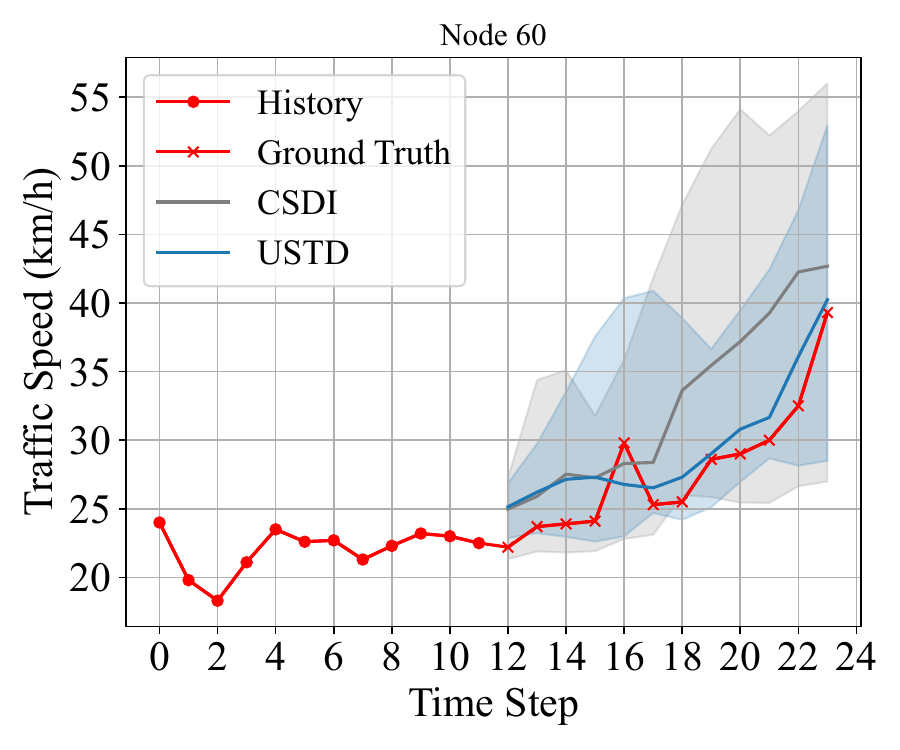}
\end{minipage}%
}

\caption{Comparison of forecasting results between USTD and CSDI on the PEMS-BAY dataset.}
\label{fig:visbay}
\end{figure*}

\section{Dataset Descriptions}
\label{app:dataset}
% We provide detailed descriptions of the datasets here.
\begin{itemize}[leftmargin=*]
    \item \emph{PEMS-03}~\cite{song2020spatial}: Traffic flow data in San Francisco Bay Area. The duration spans from Spe. 1 - Nov. 30, 2018.
    \item \emph{PEMS-BAY}~\cite{song2020spatial}: This dataset records traffic speed data of the Bay area from 325 sensors. The data frequency is 5 minutes and the data ranges from Jan. 1 - Jun. 30, 2017.
    \item \emph{AIR-BJ}~\cite{zheng2015forecasting}: An air quality dataset from 36 stations in Beijing. 
    We consider PM2.5, the most important measure for air quality systems. It has one year of data from May 1, 2014 - Apr. 30, 2015.
    \item \emph{AIR-GZ}~\cite{zheng2015forecasting}: The Guangzhou dataset has the same format as the Beijing dataset, collecting AQIs of Guangzhou by 42 stations. It has readings of 8,760 time steps starting from May 1, 2014. 
\end{itemize}

\section{Baselines Introduction}
\label{app:baseline}
We provide descriptions to baselines. If available, we adopt their official implementations in the experiments. As we cannot find the codes for ADAIN and KCN, we implemented them ourselves. 

\begin{itemize}[leftmargin=*]
    \item \emph{STGCN}~\cite{yu2018stgcn}: A pioneering model that employs GNNs and TCNs for spatial and temporal relations.
    \item \emph{DCRNN}~\cite{li2018dcrnn_traffic}: RNNs are utilized to capture temporal relations.
    \item \emph{STSGCN}~\cite{song2020spatial}: STSGCN aims to capture spatial and temporal dependencies simultaneously by a synchronous adjacency matrix.
    \item \emph{GMSDR}~\cite{liu2022msdr}: The model enhances its ability to model long-range spatial relations, with a new RNN introduced to take in multiple historical time steps at each recurring unit.
    \item \emph{STGODE}~\cite{fang2021spatial}: Spatial-Temporal Graph ODE leverages the capabilities of neural ordinary differential equations (ODE) frameworks.
    \item \emph{STGNCDE}~\cite{choi2022graph}: It utilizes neural controlled differential equations.
    \item \emph{ADAIN}~\cite{cheng2018neural}: The method is an early attention-based neural network for air quality inference.
    \item \emph{KCN}~\cite{appleby2020kriging}: A spatial model that uses graph neural networks to model spatial dependencies.
    \item \emph{IGNNK}~\cite{wu2021inductive}: A model for both spatial and temporal dimensions. 
    \item \emph{INCREASE}~\cite{zheng2023increase}: A graph network uses external information.
    \item \emph{MC Dropout}~\cite{wu2021quantifying}: Probabilistic MC Dropout adopts Monte Carlo dropout to improve the model's ability on uncertainty estimates.
    \item \emph{TimeGrad}~\cite{rasul2021autoregressive}: The first diffusion model for forecasting, which utilizes RNNs to encode historical information.
    \item \emph{CSDI}~\cite{tashiro2021csdi}: A transformer-based approach that encodes conditional data and generates target variables at the same time.
    \item \emph{DiffSTG}~\cite{wen2023diffstg}: A diffusion model for spatio-temporal forecasting based on a U-net architecture. 
    \item \emph{PriSTI}~\cite{liu2023pristi}: PriSTI utilizes a spatio-temporal encoding layer to capture conditional dependencies before the CSDI layer.
\end{itemize}

\section{Experiment Setups}
\label{app:setup}
\paragraph{Forecasting} We use signals of $12$ past time steps to predict the subsequent $12$ steps. The datasets are temporally partitioned into training, validation, and testing segments. 
We first pre-train the encoder on the training set. Then, we train the denoising module while still finetuning the encoder~\cite{liu2023spatio} at one-tenth the learning rate for the denoising module.
\paragraph{Kriging} Besides temporal partition, we also divide nodes into observed ones and target ones at a ratio of $N:M=2:1$, where observed nodes are utilized to predict signals of target nodes. The encoder is also finetuned when optimizing the denoising network.
We set the time window as $T=12$. Note that to tally with the forecasting task, the kriging task is conducted under the transductive setting. Thus, the two tasks can share the same pre-trained encoder.  
\paragraph{Hypermeters} The encoder comprises six layers, which can be divided into two blocks. Each block has three layers with the dilation factors of $[1,2,3]$. 
Both the encoder and decoder have a hidden size of 32. At the bottom layer of the encoder, a $1\times 1$ convolution maps the features into latent space with a channel of 64. We empirically find the graph sampling rate of 80\% yields the best performance. The masking rate is set to 75\%, following~\cite{he2022masked}. The TGA and SGA contain two-layer attention networks with a channel size of 96. We add temporal~\cite{zhou2021informer} or spatial~\cite{dwivedi2020generalization} embeddings to the attention. The diffusion embedding is also added to inform the diffusion steps~\cite{rasul2021autoregressive}. All the diffusion settings are identical with the previous methods~\cite{rasul2021autoregressive,tashiro2021csdi}. Our model is implemented by PyTorch 2.0 and evaluated on an RTX 3080 GPU. The training is performed using the Adam optimizer~\cite{kingma2014adam}, with a learning rate of $10^{-3}$. 

\section{Visualizations}
We provide more visualizations of USTD and the baselines on the datasets. Fig.~\ref{fig:vis03} shows the results of USTD and CSDI on PEMS-03, from which our model consistently outperforms CSDI regarding prediction accuracy and uncertainty estimates. We also obtain similar observations on the PEMS-BAY dataset, as shown in Fig.~\ref{fig:visbay}.

\end{document}